\documentclass{article}

\usepackage{mlmath}
\usepackage{PRIMEarxiv}

\usepackage[utf8]{inputenc} 
\usepackage[T1]{fontenc}    
\usepackage{hyperref}       
\usepackage{url}            
\usepackage{booktabs,multirow}       
\usepackage{amsfonts}       
\usepackage{nicefrac}       
\usepackage{microtype}      
\usepackage{xcolor}         

\usepackage{times}
\usepackage{graphicx}
\usepackage{subfigure}
\usepackage{caption}
\usepackage{amsmath,amsfonts,amssymb,mathtools,amsthm}
\usepackage{makecell}

\DeclareMathOperator*{\argmin}{arg\,min}

\newtheorem{thm}{Theorem}
\newtheorem{cor}{Corollary}

\newtheorem*{prop*}{Proposition}
\newtheorem{prop}{Proposition}
\newtheorem{rmk}{Remark}

\newtheorem{theorem}{Theorem}
\newtheorem*{theorem*}{Theorem}
\newtheorem{lemma}{Lemma}
\newtheorem*{lemma*}{Lemma}

\newtheorem*{property*}{Property}
\newtheorem*{remark}{Remark}
\newtheorem{definition}{Definition}

\newtheorem{assumption}{Assumption}
\newtheorem*{assumption*}{Assumption}

\newcommand{\rnarr}{\textnormal{narr}}
\newcommand{\rwide}{\textnormal{wide}}

\newcommand{\rrank}{\textnormal{rank}}
\newcommand{\rdim}{\textnormal{dim}}
\newcommand{\rspan}{\textnormal{span}}

\newcommand{\rNL}{\textnormal{NL}}
\newcommand{\TT}{\text{T}} 
 
\newcommand{\vep}{\varepsilon}

\title{Linear Stability Hypothesis and Rank Stratification for Nonlinear Models}
\begin{document}

\author{
Yaoyu Zhang\textsuperscript{\rm 1,4}\thanks{Corresponding author: zhyy.sjtu@sjtu.edu.cn.},
Zhongwang Zhang\textsuperscript{\rm 1}, 
Leyang Zhang\textsuperscript{\rm 2}, 
Zhiwei Bai\textsuperscript{\rm 1},
Tao Luo\textsuperscript{\rm 1,3},  
Zhi-Qin John Xu\textsuperscript{\rm 1}\thanks{Corresponding author: xuzhiqin@sjtu.edu.cn.}
\\
\textsuperscript{\rm 1}  School of Mathematical Sciences, Institute of Natural Sciences, MOE-LSC \\ 
and Qing Yuan Research Institute, Shanghai Jiao Tong University \\
\textsuperscript{\rm 2} Department of Mathematics, University of Illinois Urbana-Champaign\\
\textsuperscript{\rm 3} CMA-Shanghai\\
\textsuperscript{\rm 4} Shanghai Center for Brain Science and Brain-Inspired Technology\\
}

\maketitle

\begin{abstract}
Models with nonlinear architectures/parameterizations such as deep neural networks (DNNs) are well known for their mysteriously good generalization performance at overparameterization. In this work, we tackle this mystery from a novel perspective focusing on the transition of the target recovery/fitting accuracy as a function of the training data size. We propose a rank stratification for general nonlinear models to uncover a model rank as an ``effective size of parameters'' for each function in the function space of the corresponding model. Moreover, we establish a linear stability theory proving that a target function almost surely becomes linearly stable when the training data size equals its model rank. Supported by our experiments, we propose a linear stability hypothesis that linearly stable functions are preferred by nonlinear training. By these results, model rank of a target function predicts a minimal training data size for its successful recovery. Specifically for the matrix factorization model and DNNs of fully-connected or convolutional architectures, our rank stratification shows that the model rank for specific target functions can be much lower than the size of model parameters. This result predicts the target recovery capability even at heavy overparameterization for these nonlinear models as demonstrated quantitatively by our experiments. Overall, our work provides a unified framework with quantitative prediction power to understand the mysterious target recovery behavior at overparameterization for general nonlinear models.

\end{abstract}

\section{Introduction}
How many data points are needed for a model to recover a target function is a basic yet fundamental problem for the theoretical understanding of model fitting. For example, in linear regression, a linear target function in general can be recovered when the training data size $n$ is no less than the model parameter size $m$. Similarly, in band-limited signal recovery, we have the Nyquist-Shannon sampling theorem stating that a periodic signal with no higher frequency than $f$ (with $m=2f$ coefficients) can be exactly recovered from $n\geqslant2f$ uniformly sampled points \cite{shannon1984communication}. Above results indicate a phase transition of the target recovery accuracy as a function of the training data size at $n=m$. Then $n\geqslant m$ is often referred to as the overdetermined/underparameterized regime whereas $n< m$ is often referred to as the underdetermined/overparameterized regime. Traditional learning theory suggests that a model in the overparameterized regime is likely to overfit the data \cite{vapnik1998adaptive,bartlett2002rademacher}, thus fails to explain why overparameterized deep neural networks often generalize well in practice \cite{breiman1995reflections,zhang2016understanding}.

Motivated by the success of deep neural networks (DNNs), there emerges a trend to study general models with nonlinear architectures/parameterizations for target recovery, e.g., linear models with deep parameterization \cite{woodworth2020kernel}, deep matrix factorization models \cite{gunasekar2017implicit,li2018algorithmic,arora2019implicit}, deep linear networks \cite{saxe2013exact,lampinen2018an,gunasekar2018implicit}. It has been demonstrated that these nonlinear models are capable of recovering target functions even at heavy overparameterization. In this work, we refer to this phenomenon as the recovery mystery of nonlinear models. To understand this mystery, a popular approach is to analyze in detail the training dynamics of these nonlinear models on a case-by-case basis in order to uncover the underlying implicit bias for each nonlinear model, such as the low-rank bias in deep matrix factorization models  \cite{gunasekar2017implicit,li2018algorithmic,arora2019implicit} and low-frequency bias in deep neural networks \cite{xu_training_2018,xu2019frequency,zhang2021linear}. Following this approach, many works advance our understanding about the recovery mystery for certain nonlinear models. However, this approach encounters huge difficulty in quantitatively analyzing deep neural networks. In addition, it fails to provide a unified mechanism underlying the recovery mystery for general nonlinear models.

In this work, we take a novel approach to this recovery mystery. Specifically, we move the focus from the detailed training dynamics in previous implicit bias studies to a general macroscopic behavior---transition of the target recovery accuracy as a function of the training data size.
Our study uncovers a new quasi-determined regime at overparameterization $n<m$ in which a given model is capable of recovering a given target function. This quasi-determined regime is determined by the linear stability  for recovery of the target function (different from the numerical linear stability in Refs. \cite{wu2018sgd,mulayoff2021implicit}). Supported by experiments, we propose a linear stability hypothesis that linearly stable interpolations are preferred by nonlinear training. Importantly, by proposing a rank stratification and establishing the linear stability theory, our work show that many long standing open problems in nonlinear model fitting reduce to the validity of this hypothesis. For example, the cause of target recovery at overparameterization, the effective size of parameters and the implicit bias for a nonlinear model, as well as the advantage of the general layer-based architecture and the superiority of the convolutional architecture specifically for neural networks. Specifically, for any nonlinear model, our rank stratification uncovers a model rank for each target function in the model function space, which quantifies the data size needed for its linear stability. Therefore, our rank stratification is a powerful tool to obtain quantitative predictions about the data size needed to recover a target function in any nonlinear model. These predictions are numerically demonstrated for matrix factorization models, two-layer tanh-NNs of a fully-connected or convolutional architecture, and remain to be demonstrated for various other models.

Our linear stability hypothesis, rank stratification and linear stability theory get inspiration from experiments, predict experiments and are supported by experiments. In Section \ref{sec:LSH}, we show how we obtain the linear stability hypothesis from the experimental observation of a simple nonlinear model. To understand the condition of the linear stability, we propose a rank stratification for general nonlinear models in Section \ref{sec:rank_analysis}. Moreover, we demonstrate that the model rank of a target function obtained by rank stratification can exactly match with the transition of recovery in experiments, thus well serving as an ``effective size of parameters'' for this target function. In Section \ref{sec:LST}, we further establish the linear stability theory based on the rank stratification, which uncovers a quasi-determined regime at overparameterization with target recovery capability. In Section \ref{sec:DNNs}, we present the rank hierarchies obtained from rank stratification for NNs of different architectures. Our analysis quantifies the superiority of CNNs to fully-connected NNs for the CNN functions as further demonstrated by our experiments.

\section{Linear stability hypothesis}\label{sec:LSH}
\begin{figure}[htbp]
	\centering
 	\subfigure[$f_{\rL}(\vx;\vtheta)$]{\includegraphics[width=0.4\textwidth]{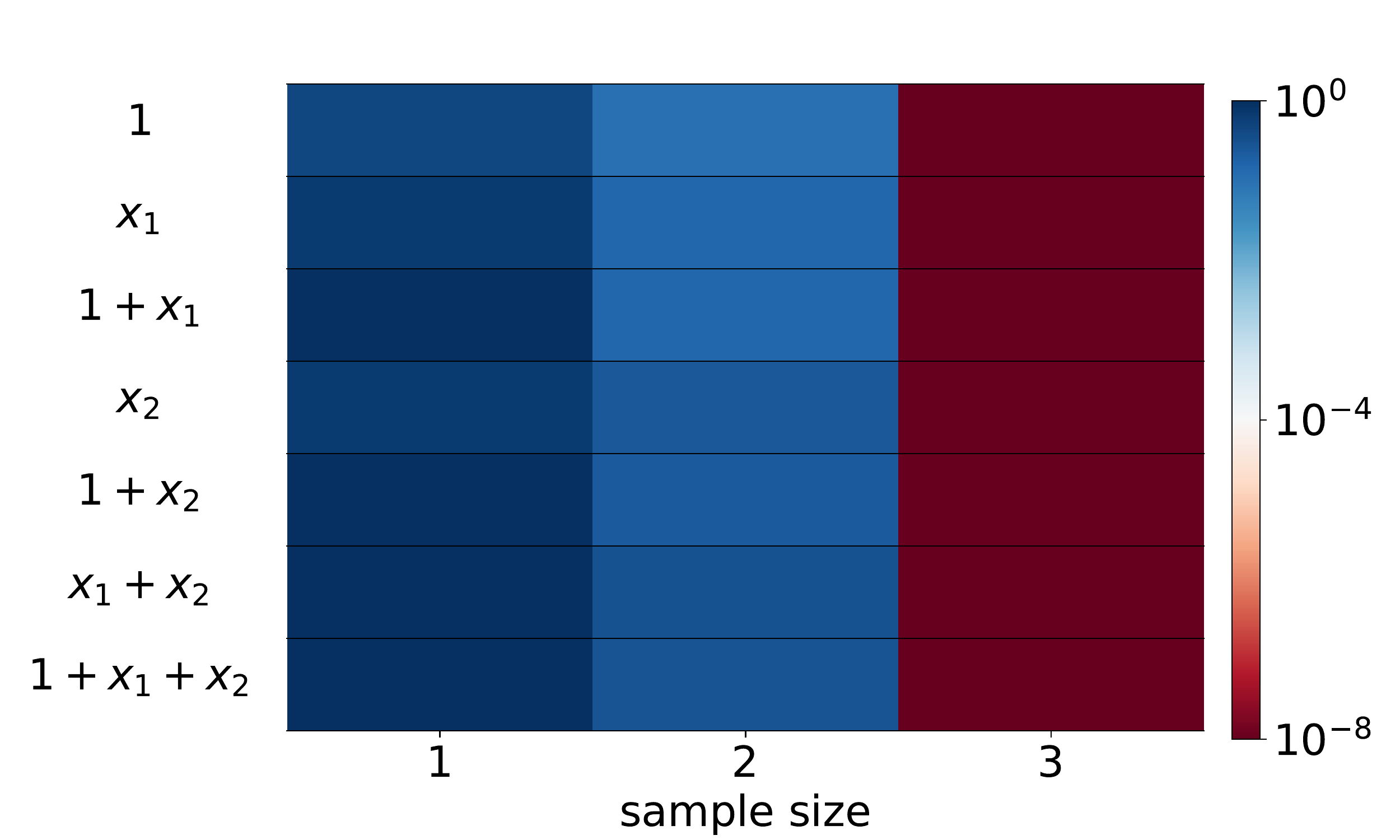}}
	\subfigure[$f_{\mathrm{NL}}(\vx;\vtheta)$]{\includegraphics[width=0.4\textwidth]{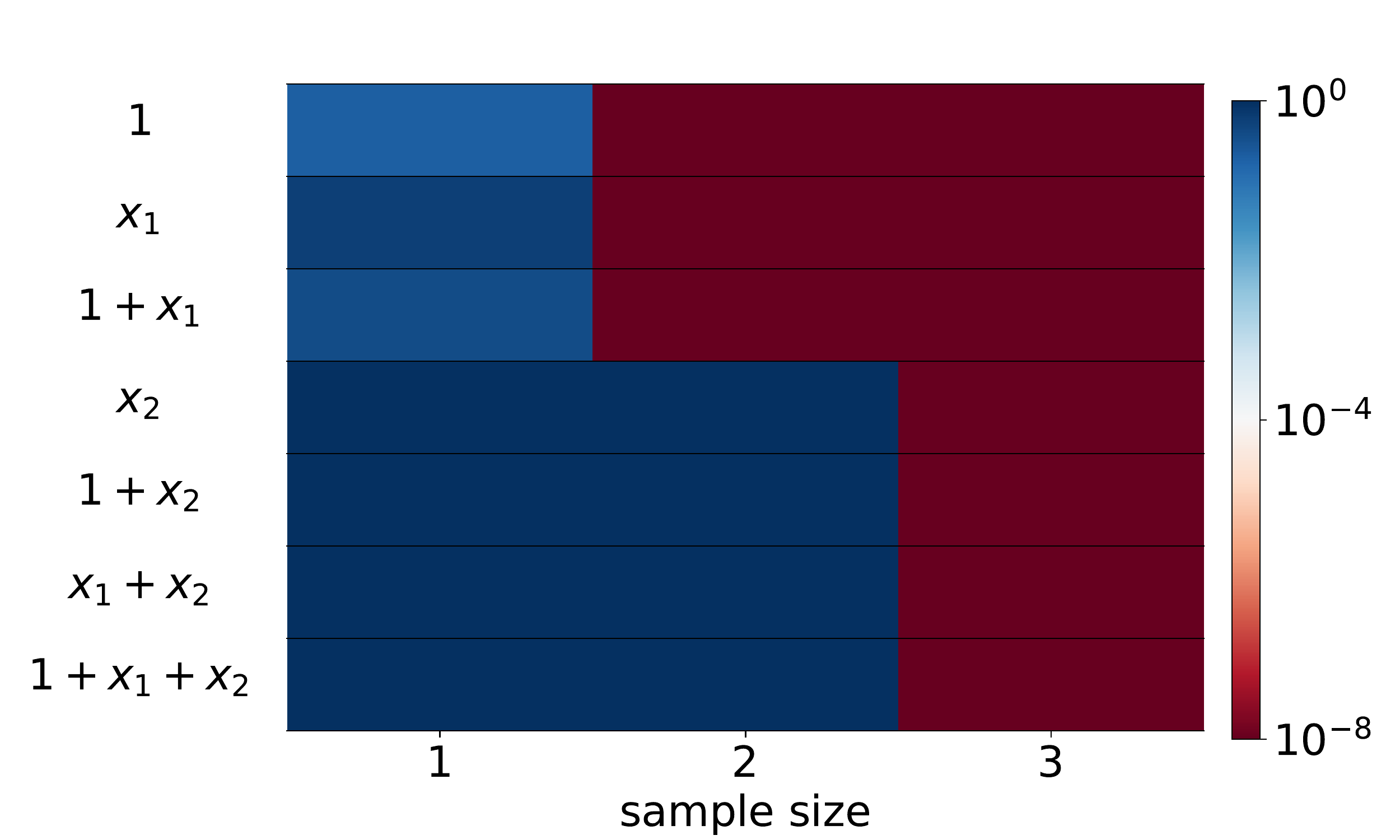}}

  \caption{Average test error (color) vs. the number of samples (abscissa) for different target functions (ordinate). For all experiments, network parameters are initialized with a normal distribution with mean $0$ and variance $10^{-8}$, and trained with full-batch GD with a learning rate of $0.01$. The training ends when the training error is less than $10^{-9}$. Each test error is  averaged over $100$ trials with random initialization.} \label{fig:random_feature_bar}
\end{figure}

Unlike linear regression, many nonlinear models (nonlinear in parameters) like DNNs are capable of accurately recovering certain target functions at overparameterization. As an example, we present the generalization accuracy of gradient descent training in recovering/fitting different target functions with various sample sizes for the following two models in Fig. \ref{fig:random_feature_bar}. One is a simple nonlinear model $f_{\mathrm{NL}}(\vx;\vtheta)=\theta_0+\theta_1 x_1+\theta_2\theta_3 x_2$ with input $\vx=[x_1,x_2]^\TT$ and parameter $\vtheta=[\theta_0,\theta_1,\theta_2,\theta_3]^\TT$. The other is its linear counterpart $f_{\rL}(\vx;\vtheta)=\theta_0+\theta_1 x_1+\theta_2 x_2$ with $\vx=[x_1,x_2]^\TT$ and $\vtheta=[\theta_0,\theta_1,\theta_2]^\TT$. These two models share the same model function space $\fF_{\rNL}=\fF_{\rL}=\{a_0+a_1x_1+a_2x_2|a_0,a_1,a_2\in\sR\}$, however clearly differ in their target recovery performance. In Fig. \ref{fig:random_feature_bar}(a), the linear model fails to recover all target functions with  training data size less than $3$ as predicted by the theory of linear regression. Surprisingly, as shown in Fig. \ref{fig:random_feature_bar}(b), the nonlinear model $\fF_{\rNL}$ accurately recovers $1$, $x_1$ and $1+x_1$ with only $2$ data points less than both its parameter size $4$ and the dimension of function space $3$. This experiment again confirms the long standing mystery that nonlinear models in general are capable of recovering certain target functions at overparameterization. Remark that, though model $\fF_{\rNL}$ is very simple, it is not easy to analyze its nonlinear training dynamics. In this situation, we take a novel approach to understand this recovery/generalization mystery by proposing the following question: \textit{When is it possible to distinguish a target minimizer (based on certain local property) from infinitely many other global minimizers at overparameterization?} Note that a target minimizer is a global minimizer whose output function equals the target function. 

To answer this question, we get inspiration from the following observation. With $2$ training data points, model $f_{\mathrm{NL}}$ finds minimizers close to $[1,0,0,0]^\TT$, $[0,1,0,0]^\TT$, and $[1,1,0,0]^\TT$ in fitting $1$, $x_1$, and $1+x_1$, respectively, as shown in Table \ref{para-table}. By looking into the tangent function space $\fT_{\vtheta}=\rspan\left\{\partial_{\theta_i} f(\cdot;\vtheta)\right\}_{i=1}^M=\rspan\{1,x_1,\theta_3 x_2,\theta_2 x_2\}$ of all the global minimizers, we notice that $\fT_{\vtheta}$ is $2$-d at $[1,0,0,0]^\TT$, $[0,1,0,0]^\TT$, and $[1,1,0,0]^\TT$, whereas $3$-d at all the other global minimizers. Remark that, given $n\geqslant r$ training data points, a global minimizer $\vtheta^*$ with a $r$-d tangent function space possesses a special local property that the corresponding function $f(\cdot;\vtheta^*)$ can be uniquely recovered in the tangent function hyperplane $\widetilde{\fT}_{\vtheta^*}=f(\cdot;\vtheta^*)+\fT_{\vtheta^*}$. For example, given $2$ different training data points, target function $1+x_1$ can be uniquely recovered in $\widetilde{\fT}_{\vtheta^*}=\{a_0+a_1x_1|a_0,a_1\in\sR\}$ for $\vtheta^*=[1,1,0,0]^\TT$. In this paper, we refer to this special local property 
as the linear stability for recovery or simply the linear stability (see Definition \ref{def:LS}). Then the output functions of these linearly stable global minimizers are referred to as the linearly stable functions. Inspired by the above observation, we propose the following \textit{linear stability hypothesis} for general nonlinear models that:

\textit{Linearly stable global minimizers are more likely to be selected by a nonlinear training process.}

Clearly, linearly stable functions are more likely to be learned by our hypothesis.
Remark that a sufficiently nonlinear training process is important for recovering a linearly stable target function in practice (an example will be shown in Fig. \ref{fig:diff_ini}). By our hypothesis, the linear stability of a target function is the key for its successful recovery. This can be analyzed rigorously for general nonlinear models as detailed in the latter sections. In the following, we first propose a rank stratification for general models to uncover a minimal training data size needed for a target function to be linearly stable. 

\begin{table}[htb]
\caption{Recovered parameter values with standard deviations for $\fF_{\rNL}$ over 100 trials for experiments of the first $3$ rows in Fig. \ref{fig:random_feature_bar}(b).}\label{para-table}
\centering
\renewcommand{\arraystretch}{1.5} 
\begin{tabular}{|c|c|c|c|}
\hline
\multirow{2}{*}{parameter}          & \multicolumn{3}{c|}{target function}                                         \\ 
  \cline{2-4}
        &$1$  & $x_1$& $1+x_1$                                                \\ \hline
$\theta_0$            & $1.0 \times 10^{0} \pm 9.8 \times 10^{-5}$ & $4.4 \times 10^{-7} \pm 4.5 \times 10^{-5}$& $1.0 \times 10^{0} \pm 2.7 \times 10^{-5}$                       \\ \hline
$\theta_1$             & $4.9 \times 10^{-5} \pm 3.0 \times 10^{-4}$ & $1.0 \times 10^{0} \pm 5.5 \times 10^{-5}$& $1.0 \times 10^{0} \pm 7.1 \times 10^{-5}$                                                  \\ \hline
$\theta_2$             & $4.0 \times 10^{-4} \pm 1.8 \times 10^{-3}$ & $1.2 \times 10^{-4} \pm 1.4 \times 10^{-3}$& $1.9 \times 10^{-4} \pm 1.4 \times 10^{-3}$                                                       \\ \hline

$\theta_3$             & $1.1 \times 10^{-4} \pm 1.9 \times 10^{-3}$ & $6.0 \times 10^{-6} \pm 1.5 \times 10^{-3}$& $2.3 \times 10^{4} \pm 1.4 \times 10^{-3}$                                                      \\ \hline
\end{tabular}
\end{table}

\section{Rank stratification}\label{sec:rank_analysis}
For a linear model $\sum_{i=1}^m\theta_i\phi_i(\vx)$ with basis functions $\{\phi_i(\cdot)\}_{i=1}^m$, it is well known that any target function in its function space can be stably recovered when the size of training data $\{(\vx_j,y_j)\}_{j=1}^n$ is no less than its effective size of parameters (or effective degrees of freedom) $\rdim(\rspan\{\phi_i(\cdot)\}_{i=1}^m)$. Remark that the above intuitive argument requires a mild assumption on data that $\rrank(\vphi(\mX))=\rdim(\rspan\{\phi_i(\cdot)\}_{i=1}^m)$, where $[\vphi(\mX)]_{i,j}=\phi_i(\vx_j)$ for $i\in[m],j\in[n]$. Therefore, for the linear model $\theta_0+\theta_1 x_1+\theta_2 x_2$ with $3$ effective parameters, we observe target recovery at $n=3$ as shown in Fig. \ref{fig:random_feature_bar}. 
Above result for a linear model can be directly applied to understand the stability of recovery in the tangent function hyperplane of any parameter point for a nonlinear model. For any nonlinear model $f_{\vtheta}=f(\cdot;\vtheta)$, $\widetilde{\fT}_{\vtheta^*}=\{f(\cdot;\vtheta^*)+\va^\TT\nabla_{\vtheta}f(\cdot;\vtheta^*)|\va\in\sR^M\}$ at $\vtheta^*\in\sR^M$ has $\rdim\left(\rspan\left\{\partial_{\theta_i} f(\cdot;\vtheta^*)\right\}_{i=1}^M\right)$ effective parameters. In $\widetilde{\fT}_{\vtheta^*}$, $f(\cdot;\vtheta^*)$ in general can be stably recovered when $n\geqslant \rdim\left(\rspan\left\{\partial_{\theta_i} f(\cdot;\vtheta^*)\right\}_{i=1}^M\right)$. A special feature of many nonlinear models is that the effective size of parameters changes over the parameter space.
In this work, we formally define this effective size of parameters as the model rank of $\vtheta^*\in\sR^M$ with respect to the model $f_{\vtheta}$, i.e., 
\begin{equation}\label{rank_p}
\rrank_{f_{\vtheta}}(\vtheta^*):=\rdim\left(\rspan\left\{\partial_{\theta_i} f(\cdot;\vtheta^*)\right\}_{i=1}^M\right).
\end{equation}
This definition of model rank is consistent with the definition of rank in differential topology. Remark that, notation $\rrank(\cdot)$ without a subscript refers to the matrix rank by default in our work.

Understanding the distribution of the model rank over the parameter space is the first step for a linear stability analysis. As an example, for the nonlinear model $f_{\vtheta}(\vx)=\theta_0+\theta_1 x_1+\theta_2\theta_3 x_2$, the model rank at any point $\vtheta^*\in \sR^{4}$ is adaptive as follows 
\begin{equation}
\rrank_{f_{\vtheta}}(\vtheta^*)=\rdim\left(\rspan\left\{1,x_1,\theta^{*}_3 x_2, \theta^{*}_2 x_2\right\}\right)=\left\{
\begin{aligned}
2, &\ \ \theta^*_2=\theta^{*}_3=0, \\
3, &\ \ \rm{others}. \\
\end{aligned} \label{def:rank_theta}
\right.    
\end{equation}

To consider the linear stability for a target function $f^*$, one must note that $\fM_{f^*}:=\{\vtheta|f(\cdot;\vtheta)=f^*;\vtheta\in\sR^M\}$ referred to as the target stratifold in this work is a disjoint union of manifolds with different dimensions and model ranks.
For example, the target stratifold for $1+x_1$ is $\{\vtheta|\theta_0=1,\theta_1=1,\theta_2\theta_3=0\}$, on which rank-$2$ is attained only at $\vtheta=[1,1,0,0]^\TT$ and rank-$3$ is attained elsewhere. When $n\geqslant 2$, target function $1+x_1$ is stable for recovery at the tangent function hyperplane of $\vtheta=[1,1,0,0]^\TT$ under a mild assumption. Then, our linear stability hypothesis predicts that $1+x_1$ is likely to be recovered through training as numerically demonstrated in Fig. \ref{fig:random_feature_bar}. 

To quantify the minimal data size needed to recover a target function $f^*$ in the model function space $\fF_{f_{\vtheta}}:=\{f(\cdot;\vtheta)|\vtheta\in\sR^M\}$, we formally define its model rank as 
\begin{equation}
    \rrank_{f_{\vtheta}}(f^*):=\min_{\vtheta'\in\fM_{f^*}}\rrank_{f_{\vtheta}}(\vtheta'),\label{rank_f*}
\end{equation}
with a slight misuse of the notion $\rrank_{f_{\vtheta}}(\cdot)$ to return the model rank for a function input. Then, the second step for a linear stability analysis is to stratify $\fF_{f_{\vtheta}}$ into function sets of different model ranks from low to high, which forms a rank hierarchy. For the linear model $\theta_0+\theta_1 x_1+\theta_2 x_2$ with a constant rank, its whole function space is rank-$3$. For the nonlinear model $\theta_0+\theta_1 x_1+\theta_2 \theta_3 x_2$, the rank hierarchy is as follows, 
\begin{equation}\label{eq:rank_NL}
\rrank_{f_{\vtheta}}(f^*)=\left\{
\begin{aligned}
2, &\ \ f^*\in\{a_0+a_1x_1|a_0,a_1\in\sR\}, \\
3, &\ \ f^*\in\{a_0+a_1x_1+a_2x_2|a_2\neq 0,a_0,a_1,a_2\in\sR\}. \\
\end{aligned}
\right.
\end{equation}
This result shows that $1$, $x_1$ and $1+x_1$ are rank-$2$, whereas all the other functions with nonzero coefficients in $x_2$ in Fig. \ref{fig:random_feature_bar} are rank-$3$. Then, our linear stability hypothesis predicts recovery with $2$ data points for $1$, $x_1$ and $1+x_1$, and $3$ data points for other function through nonlinear training. Clearly, this prediction perfectly matches with the experimental results in Fig. \ref{fig:random_feature_bar}.

In general, for a differentiable model $f_{\vtheta}$ with $M$ parameters, the standard procedure of rank stratification is comprised of the following two steps: (1) stratify the parameter space into different model rank levels to obtain the rank hierarchy over the parameter space; (2) stratify the model function space into different model rank levels to obtain the rank hierarchy over the model function space. Remark that, the difficulty of rank stratification depends on the complexity of model architecture as shown in the following sections. 

The rank stratification proposed above uncovers that different functions in the function space of a nonlinear model may have different effective sizes of parameters. 
Clearly, even when two models share the same model function space, different parameterization/architecture can lead to very different hierarchies as demonstrated by the above comparison between $f_{\rL}$ and $f_{\rNL}$. By our linear stability hypothesis, this rank hierarchy indicates an implicit bias towards low model rank functions over the model function space as detailed later in Section \ref{sec:LST}.
Overall, the proposed rank stratification is a powerful tool that could explicitly uncover an architecture-specific implicit bias of a nonlinear model.
In the following subsection, we present the rank hierarchy obtained by the rank stratification in a table for a nonlinear matrix factorization model, and demonstrate the relation predicted by our hypothesis between the model rank of a target function and its experimental transition of the target recovery accuracy.

\subsection{Matrix factorization model: rank hierarchy matches with the transition of target recovery}
\begin{figure}[htbp]
	\centering
	\includegraphics[width=0.5\textwidth]{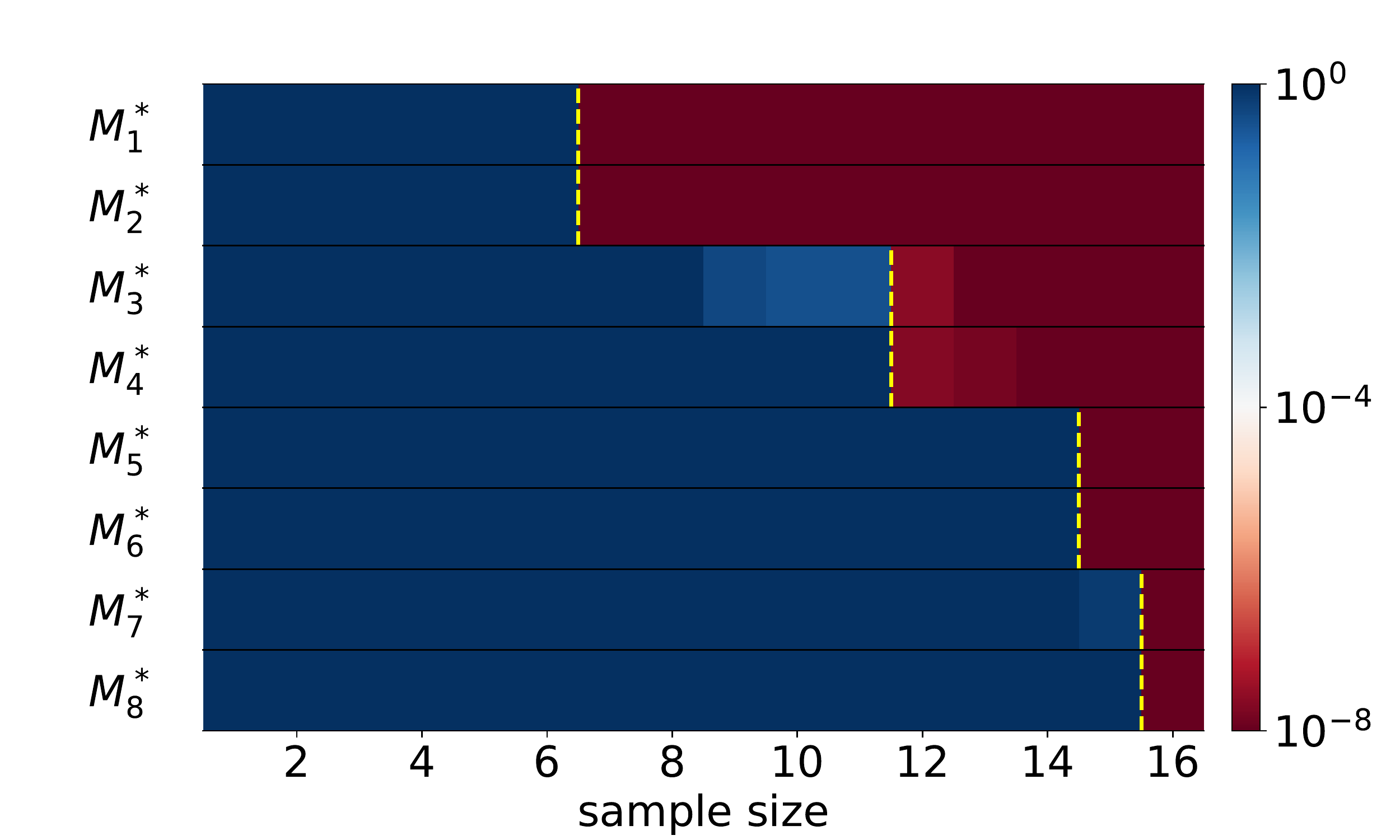}
	
  \caption{Average test error (color) vs. the number of samples (abscissa) for different target functions (ordinate). The yellow dashed line for each row indicates the transition when the size of the observed entries equals to the model rank of the target. Different rows represent different target matrices (see Appendix Section \ref{appsec:exp} for details). In particular, $\rrank(\mM_{2k-1}^*)=\rrank(\mM_{2k}^*)=k$ for $k=1,2,3,4$. For all experiments, the weights are initialized with a normal distribution with mean $0$ and variance $10^{-8}$, and trained with full-batch GD with a learning rate of $0.05$. The training ends when the training error is less than $10^{-9}$. Each test error is averaged over $50$ trials with random initialization.} \label{fig:matrix_com}
\end{figure}

To demonstrate the power of the rank stratification for general nonlinear models, we consider in this section a practical nonlinear model of matrix factorization $\vf_{\vtheta} = \mA\mB$ with application in matrix completion. 
In Table~\ref{table:rank_MF}, we present its rank hierarchy obtained through rank stratification (see Appendix Section \ref{appsec:MF} for details). All elements in matrices $\mA$ and $\mB$ are trainable parameters. By Table~\ref{table:rank_MF}, a target matrix $\vf^*$ with matrix rank $1$, $2$, $3$, or $4$ possesses the model rank $7$, $12$, $15$, or $16$, respectively. It can be clearly seen from Fig. \ref{fig:matrix_com}  that these model ranks exactly match with the transition of the target recovery accuracy, i.e., the test error drops rapidly to almost $0$ when the size of the observed entries equals the model rank of the target (marked by a yellow dashed line). This result further demonstrates the importance of rank stratification for understanding the target recovery behavior of nonlinear models, and supports the validity of the linear stability hypothesis.

\begin{table}[htbp]
\centering
\renewcommand{\arraystretch}{1.5} 
\begin{tabular}{|c|c|}
\hline
 \multicolumn{2}{|c|}{model:$\vf_{\vtheta} = \mA\mB, \vtheta = (\mA, \mB), \mA, \mB \in \sR^{d\times d}$}                                         \\ \hline
$\operatorname{rank}_{\vf_{\vtheta}}(\vf^{*})$                  & \multicolumn{1}{c|}{$\vf^*$}                                                        \\ \hline
$0$                                           & \multicolumn{1}{c|}{$\boldsymbol{0}_{d\times d}$}                                       \\ \hline
$2d-1$                                        & \multicolumn{1}{c|}{$\operatorname{rank}(\vf^*) = 1$} \\ \hline
$\vdots$                                      & \multicolumn{1}{c|}{$\vdots$}                       \\ \hline
$2rd-r^2$                                     & \multicolumn{1}{c|}{$\operatorname{rank}(\vf^*) = r$} \\ \hline
$\vdots$                                      & \multicolumn{1}{c|}{$\vdots$}                        \\ \hline
$d^2$                                         & \multicolumn{1}{c|}{$\operatorname{rank}(\vf^*) = d$} \\ \hline
\end{tabular}
\vspace{5pt}
\caption{Rank hierarchy for the matrix factorization model.\label{table:rank_MF}}
\end{table}

\begin{figure}[htbp]
	\centering
 	\subfigure[use $f_{\rm NL}$ to fit, $1+x_1$]{\includegraphics[width=0.45\textwidth]{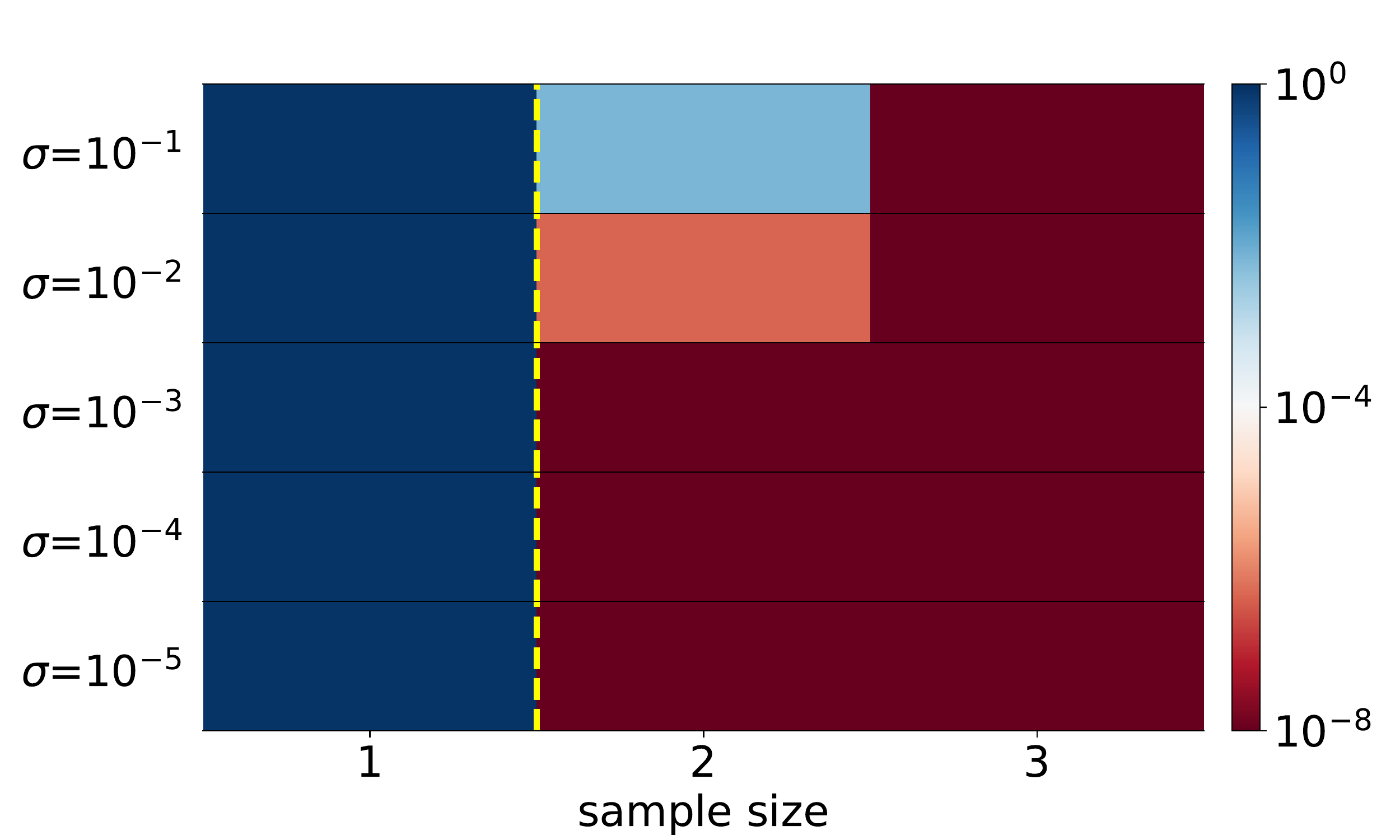}}
	\subfigure[matrix completion, rank=1]{\includegraphics[width=0.45\textwidth]{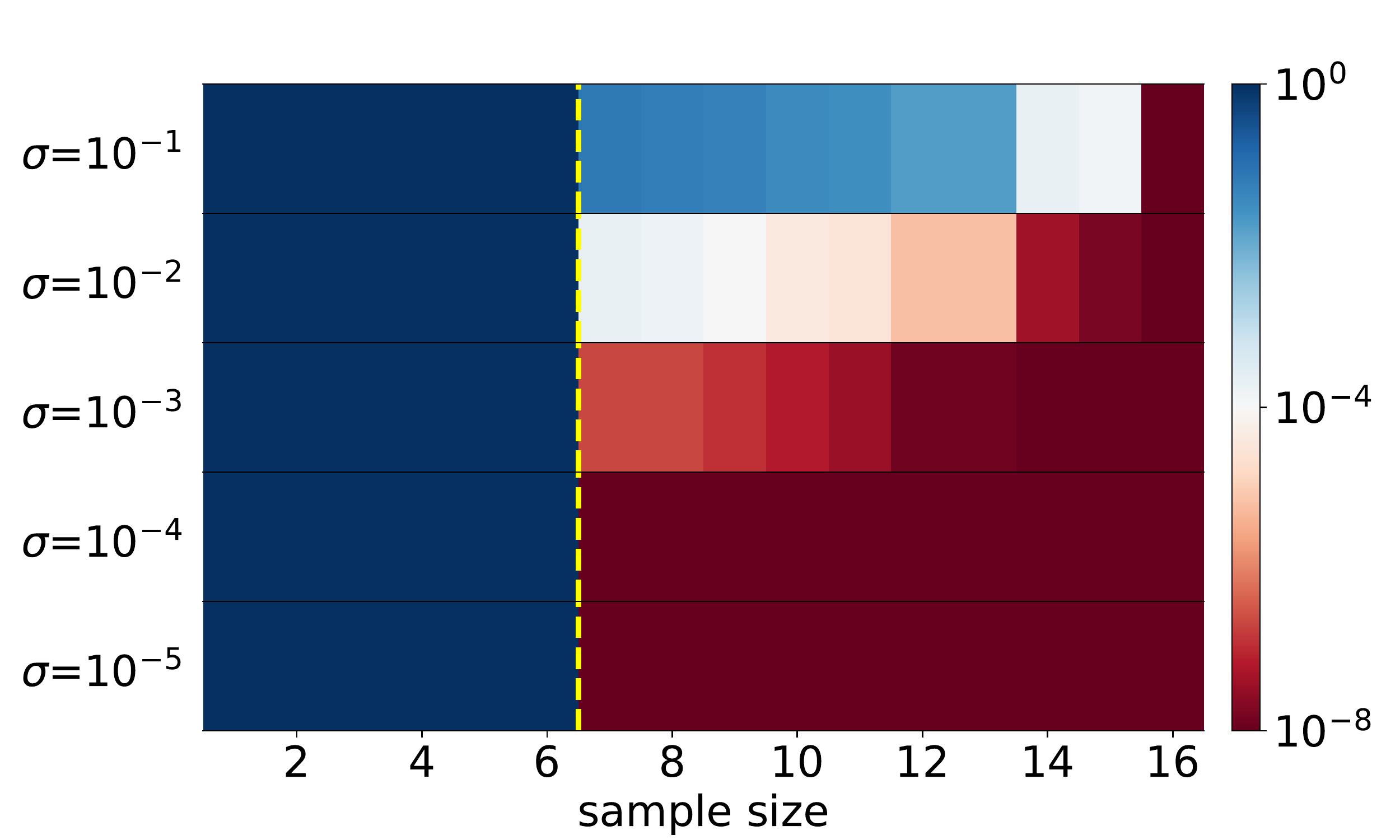}}
\caption{Average test error (color) vs. the number of samples (abscissa) over different sizes of standard deviation of initialization (ordinate). (a) Performance of fitting $1+x$ by $f_{\rm NL}(x,\theta)$. (b) Performance of completing a rank-$1$ matrix by the matrix factorization model. We use $\mM_{1}^*$ in Fig. \ref{fig:matrix_com} as our target matrix, whose specific form is given in the appendix. The yellow dashed line for each row indicates the transition when the sample size equals to the model rank of the target. For all experiments, the network is initialized with a normal distribution with mean $0$ and variance $\sigma^2$, and trained with full-batch GD with a learning rate $0.05$. Each test error is averaged over $50$ trials with random initialization. } \label{fig:diff_ini}
\end{figure} 

\subsection{Quasi-determined regime}
From above results, it is clear that there exists a new regime for nonlinear models at overparameterization where a target function can be successfully recovered. We name this regime the quasi-determined regime formally defined later in Definition \ref{def:qd_regime}.
From experiments, \textit{the quasi-determined regime covers a wide range of training data sizes from the model rank of the target function to the size of model parameters}. This adaptiveness to the target function is the key characteristic of the quasi-determined regime absent in conventional regime characterization.
Remark that, the quasi-determined regime is specific to rank-adaptive models in which model ranks are non-constant over their function spaces. In a model with a constant model rank such as a linear model, the quasi-determined regime is empty because the model rank of any function in the model function space equals the dimension of the function space above which the fitting problem becomes determined/over-determined.

As shown in Fig. \ref{fig:diff_ini}, target recovery in the quasi-determined regime is very different from that in the over-determined/underparameterized regime: (i) The success or accuracy of recovery depends on the initialization. One may need to tune the scale of initialization to a sufficiently small value, which leads to a highly nonlinear training dynamics, in order to achieve a good recovery accuracy. (ii) Increasing the size of training data above the model rank of the target function further increases the tolerance on the initialization scale and enhances the accuracy of recovery. These two properties match with the widely-observed behavior of DNNs and other nonlinear models in practice that good hyperparameter tuning and a large training data size are two important factors for an accurate target recovery at overparameterization. Therefore, it is reasonable to believe that our quasi-determined regime is relevant to the training of general nonlinear models in practice. To understand the exact relation among our rank stratification, the quasi-determined regime and the linear stability hypothesis, we establish in the following the linear stability theory for recovery for general models.

\section{Linear stability theory}\label{sec:LST}

In this section, we rigorously analyze the linear stability for general models to address when a function or a minimizer becomes linearly stable for recovery in the linearized function space, i.e., the tangent function hyperplane. By admitting the linear stability hypothesis, our results in the following yield quantitative predictions to the global target recovery behavior of nonlinear models. Remark that our linear stability hypothesis and analysis is inspired by the widely-used linear stability analysis in mathematics, which serves as a powerful tool to understand the first-order behavior of a nonlinear system. Also note that all the linear stability in our work refers to the linear stability for recovery in Definition \ref{def:LS}. It is starkly different from the commonly considered numerical linear stability for neural networks originated from the numerical discretization of a continuous training dynamics \cite{wu2018sgd,mulayoff2021implicit}.
Our analysis begins with the following formal definition of the linear stability.

\begin{definition}[linear stability for recovery\label{def:LS}]
Given any differentiable model $f_{\vtheta}$ with model function space $\fF_{f_{\vtheta}}$, loss function $\ell(\cdot,\cdot)$, and training data $S=\{(\vx_i,y_i)\}_{i=1}^n$, \\
(i) a parameter point $\vtheta^*\in\sR^M$ is linearly stable if $f(\cdot;\vtheta^*)$ is the unique solution to 
\begin{equation}
    \mathrm{min}_{f\in\widetilde{\fT}_{\vtheta^*}}\frac{1}{n}\sum_{i=1}^{n}\ell(f(\vx_i),y_i); \label{eq:minF}
\end{equation}\\
(ii) a function $f^*\in\fF_{f_{\vtheta}}$ is linearly stable if there exists a linearly stable parameter point $\vtheta'$ such that $f(\cdot;\vtheta')=f^*$.
\end{definition}
Without loss of generality, we consider $\ell(\cdot,\cdot)$ to be a continuously differentiable distance function by default and focus on studying the linear stability of the global minimizers attaining $0$ loss as well as the corresponding interpolations, i.e., functions in $\fF_{f_{\vtheta}}$ attaining $0$ loss. By the linear stability hypothesis, above formal definition of the linear stability immediately gives us a novel regime with target recovery capability defined as follows.

\begin{definition}[quasi-determined regime\label{def:qd_regime}]
Using any model $f_{\vtheta}$ to fit a target function $f^*\in\fF_{f_{\vtheta}}$ from data $S=\{(\vx_i,f^*(\vx_i))\}_{i=1}^{n}$ at overparameterization, the fitting problem is quasi-determined if $f^*$ is linearly stable for recovery.
\end{definition}

In the following, we present our theory of linear stability beginning with a necessary and sufficient condition for linear stability.

\begin{lemma}[linear stability condition, see Lemma \ref{thm:LSC_app} in Appendix for proof]\label{thm:LSC}
Given any differentiable model $f_{\vtheta}$ and training data $S=\{(\vx_i,y_i)\}_{i=1}^n$,
a global minimizer $\vtheta^*$ satisfying $f(\vx_i;\vtheta^*)=y_i$ for all $i\in[n]$ is linearly stable if and only if $\rrank_{S}(\vtheta^*)=\rrank_{f_{\vtheta}}(\vtheta^*)$, where the empirical model rank $\rrank_{S}(\vtheta^*):=\rrank(\nabla_{\vtheta}f(\mX;\vtheta^*))$.
\label{lem:lin_sta_cond}
\end{lemma}

\begin{figure}[htbp]
	\centering
	\includegraphics[width=0.6\textwidth]{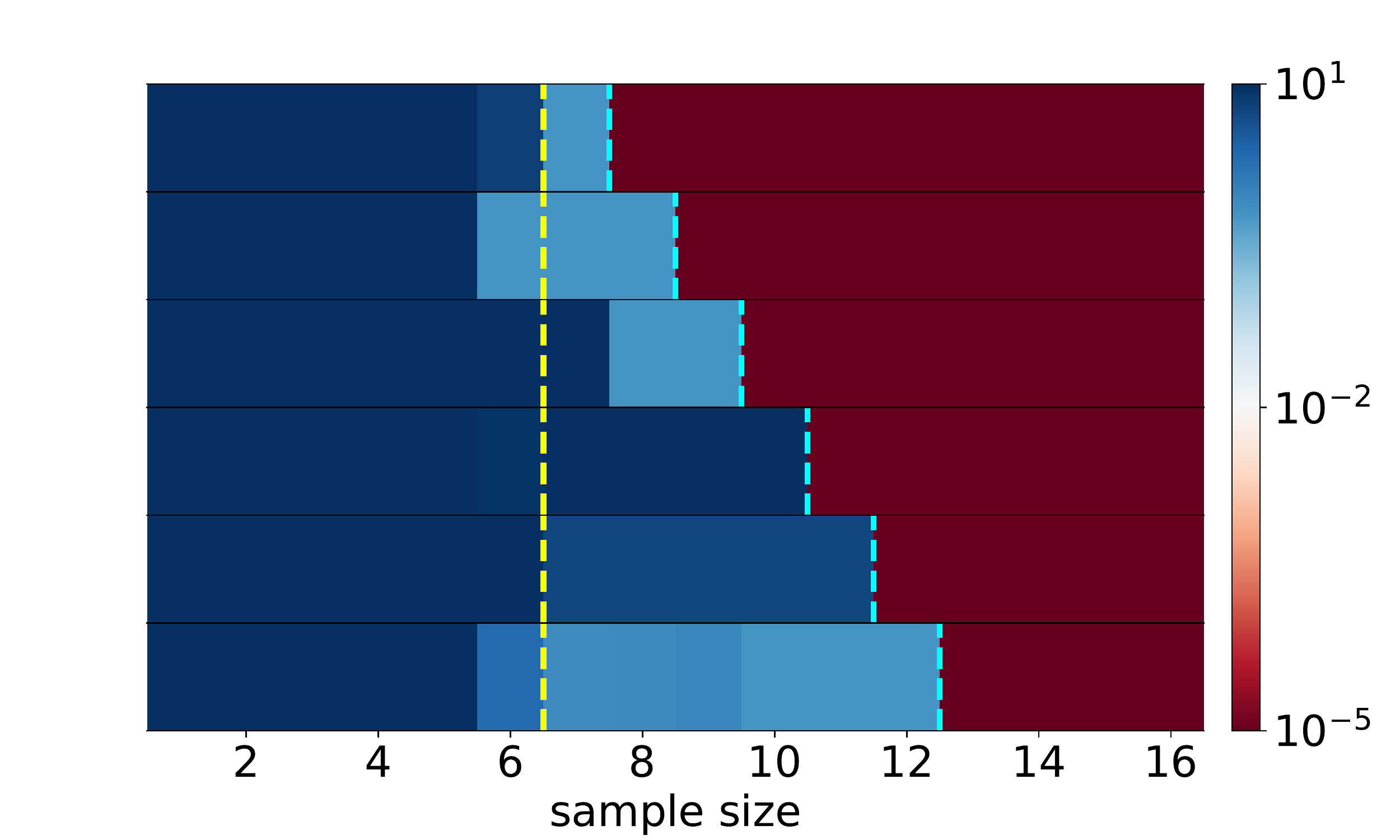}
  \caption{The relationship between the number of samples (abscissa) and the test error (color) for different sampling sequences with specifically designed orders (ordinate) to reconstruct a target matrix with matrix rank one. For a specific sampling sequence, $n_t=\mathop{\min}\limits_{n}\{n|\rrank_{S_n}(\vtheta^*)=\rrank_{f_{\vtheta}}(\vtheta^*)\}$ is indicated by cyan dashed curve. The model rank of the target matrix is $7$ indicated by the yellow dashed curve. For all experiments, the parameters are initialized by a normal distribution with mean $0$ and variance $10^{-8}$, and trained by the full-batch gradient descent with a learning rate $0.05$. Each test error in the figure is averaged over $50$ trials with random initialization.} \label{fig:matrix_com2}
\end{figure}

Note that $\nabla_{\vtheta}f(\mX;\vtheta^*)=[\nabla_{\vtheta}f(\vx_1;\vtheta^*),\cdots, \nabla_{\vtheta}f(\vx_n;\vtheta^*)]$ with $\mX:=[\vx_1,\cdots,\vx_n]$ is referred to as the empirical tangent matrix. By the above lemma, we can determine the linear stability of a target function by checking whether there exists a target minimizer satisfying this linear stability condition for the given training data.
The intuition of Lemma \ref{thm:LSC} is as follows. In the tangent function hyperplane $\widetilde{\fT}_{\vtheta^*}$, $\rrank_{S}(\vtheta^*)$ quantifies the number of independent constraints from data, and $\rrank_{f_{\vtheta}}(\vtheta^*)$ quantifies the effective size of parameters. Therefore, the linearized problem Eq. (\ref{eq:minF}) at a global minimizer $\vtheta^*$ becomes determined if and only if $\rrank_{S}(\vtheta^*)=\rrank_{f_{\vtheta}}(\vtheta^*)$. Lemma \ref{thm:LSC} implies cases in which a target function may not become linearly stable with $n=\rrank_{f_{\vtheta}}(\vtheta^*)$ training data points due to the lack of data independence. For these cases, our linear stability hypothesis predicts a transition of the target recovery accuracy later than $n=\rrank_{f_{\vtheta}}(\vtheta^*)$.
We numerically verify this prediction by the following experiments. As shown in Fig. \ref{fig:matrix_com}, the minimum sample size to recover a rank one matrix is 7. However, we can design a sample sequence $\{(i_1,j_1),(i_2,j_2),\cdots\}$ with $S_n=\{((i_k,j_k),\vf^*_{i_kj_k})\}_{k=1}^n$ such that the minimum sample size to satisfy the linear stability condition in Lemma \ref{thm:LSC} is larger than 7, that is, $n_t=\mathop{\min}\limits_{n}\{n|\rrank_{S_n}(\vtheta^*)=\rrank_{f_{\vtheta}}(\vtheta^*)\}>7$. In Fig. \ref{fig:matrix_com2}, each row indicates a specially designed sample sequence with a different $n_t$ indicated by the cyan dashed curve. Clearly, only when the linear stability condition is satisfied, i.e., sample size is no less than $n_t$, the test error drops rapidly to almost $0$. All these experiments confirm that the experimental transition of the target recovery accuracy matches with the transition of the linear stability of the target. Again, they support the validity of our linear stability hypothesis.

Lemma \ref{thm:LSC} provides the exact condition about when a global minimizer becomes linearly stable. However, checking this condition for various global minimizers is a tedious job. In practice, it is important to have a more convenient and intuitive condition. For example, a model of $m$ parameters can be recovered almost surely from $n\geqslant m$ data points for linear regression. In analogy, we find out that $f^*$ becomes linearly stable with $n\geqslant \rrank_{f_{\vtheta}}(f^*)$ data points almost surely for a model analytic with respect to its parameters by the following theorem. 

\begin{figure}[htbp]
	\centering
 	\subfigure[matrix rank]{\includegraphics[width=0.33\textwidth]{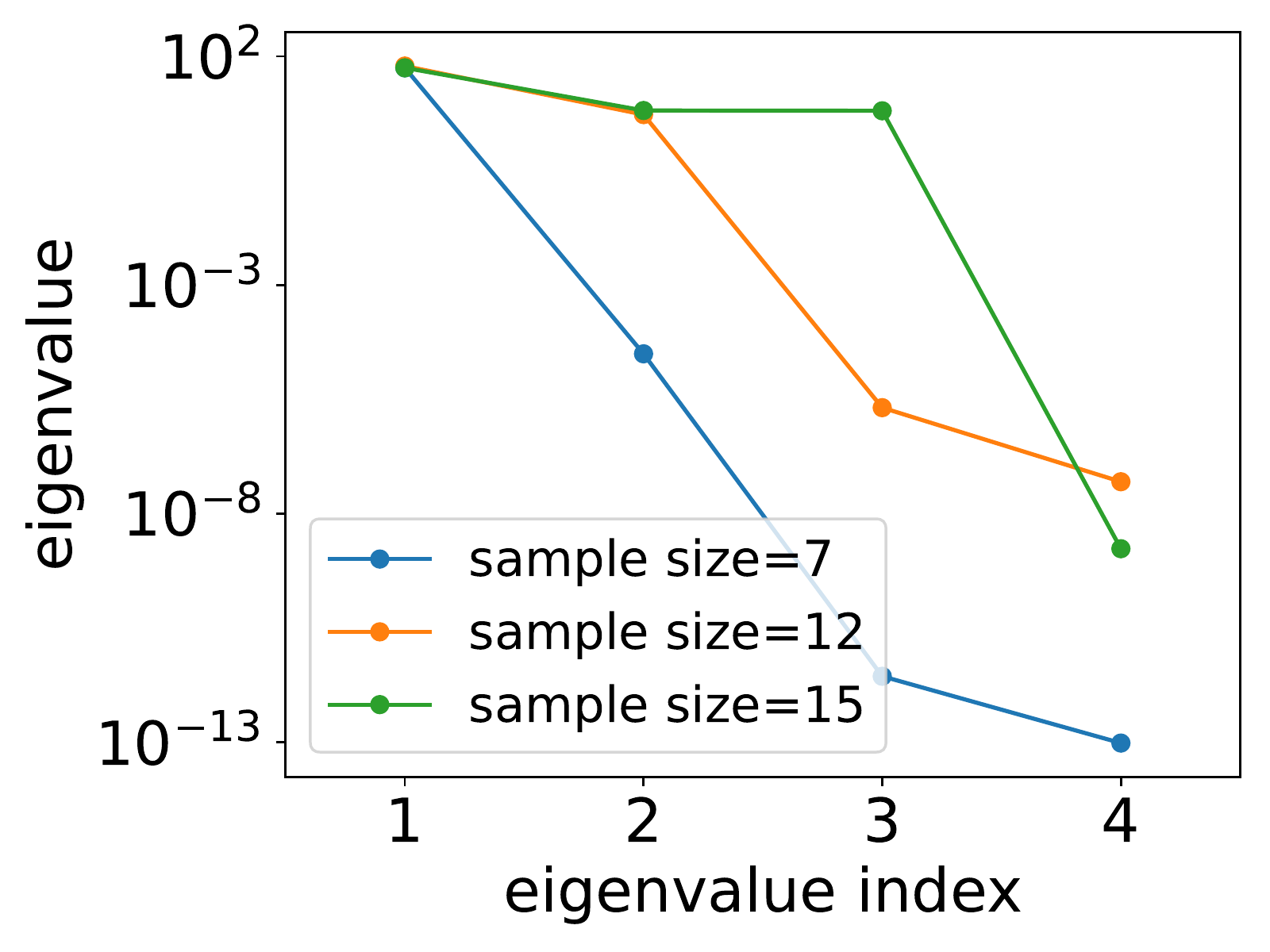}}
	\subfigure[sample size=7]{\includegraphics[width=0.33\textwidth]{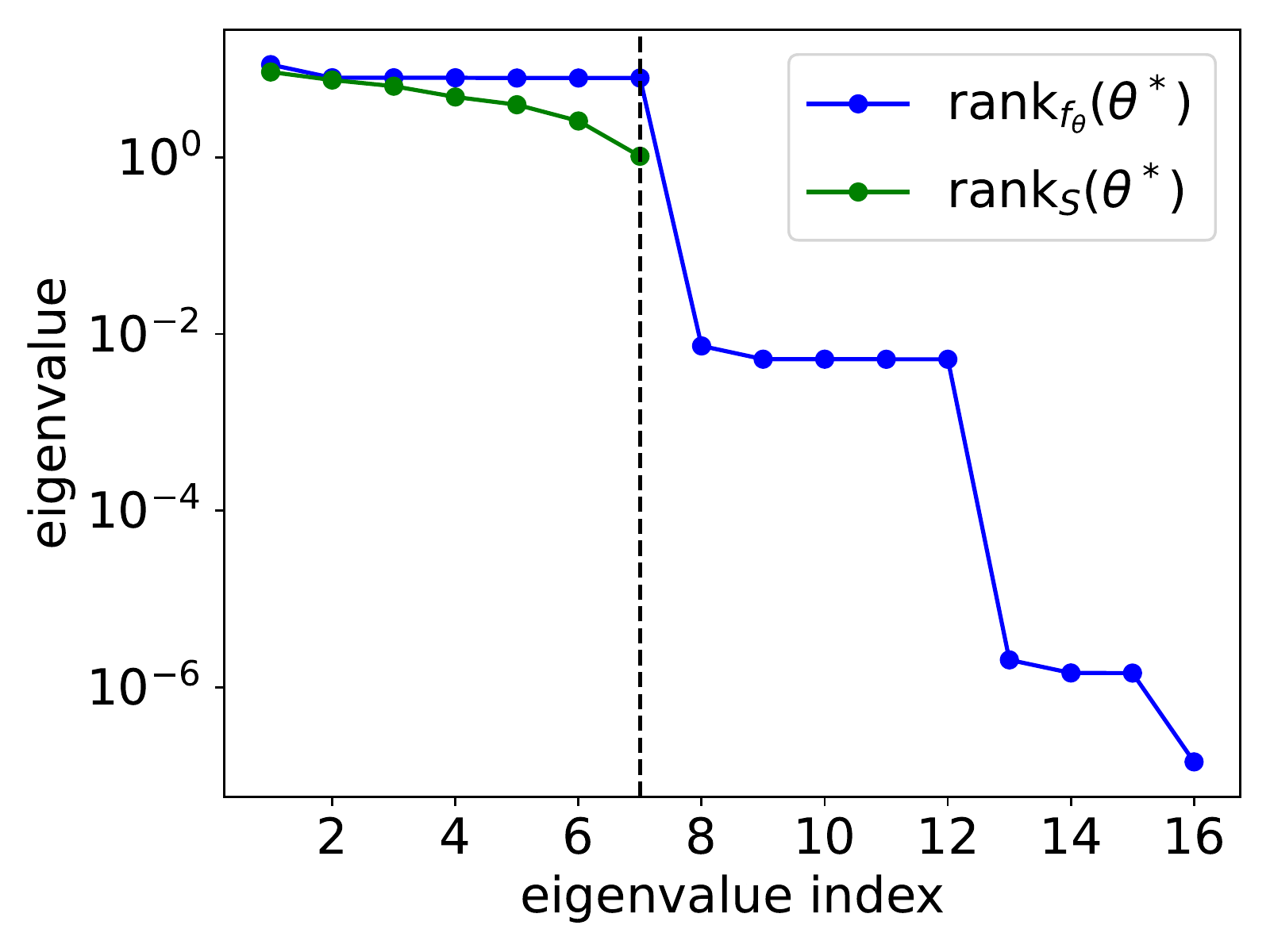}}\\
        \subfigure[sample size=12]
    {\includegraphics[width=0.33\textwidth]{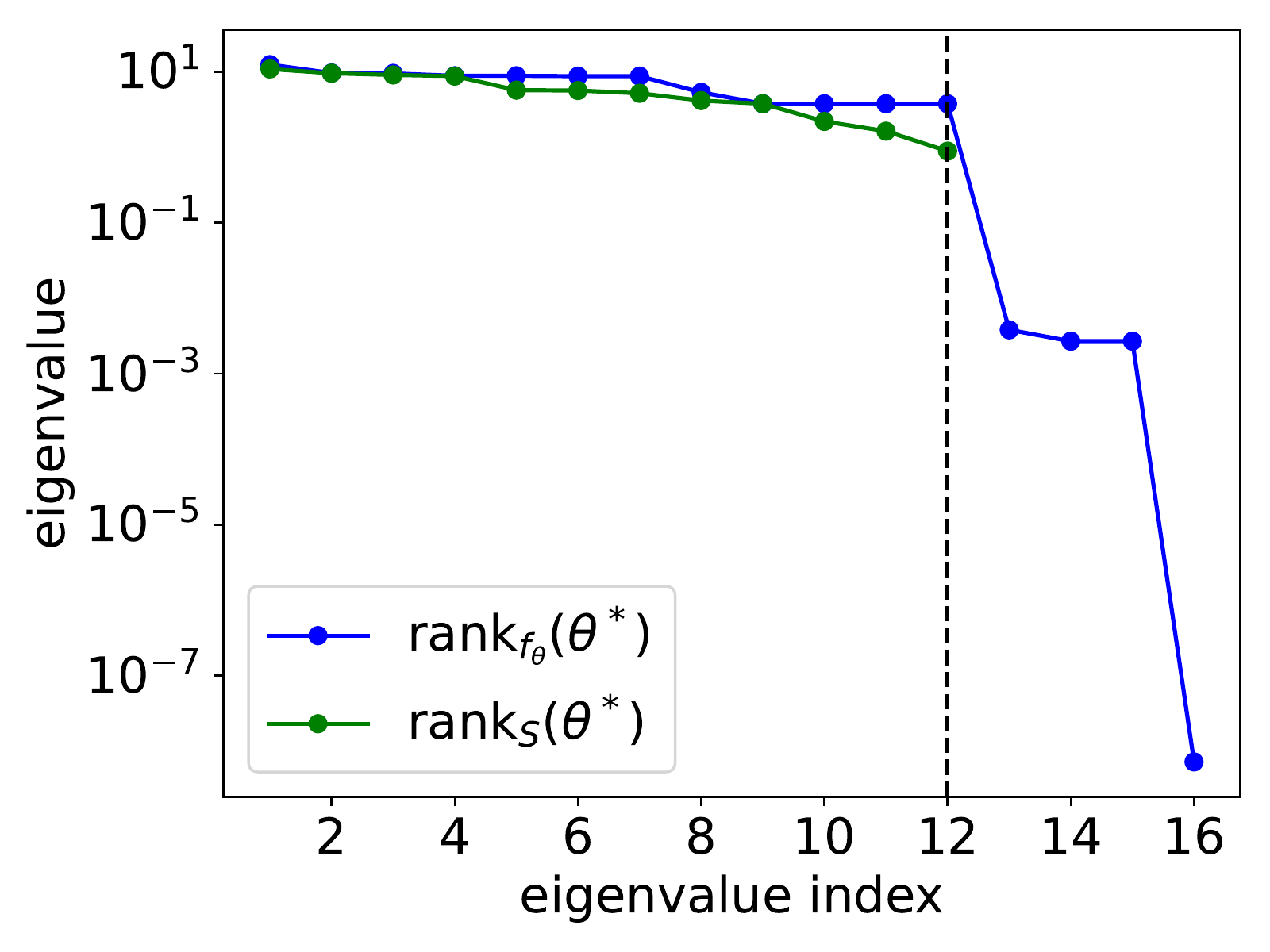}}
        \subfigure[sample size=15]
    {\includegraphics[width=0.33\textwidth]{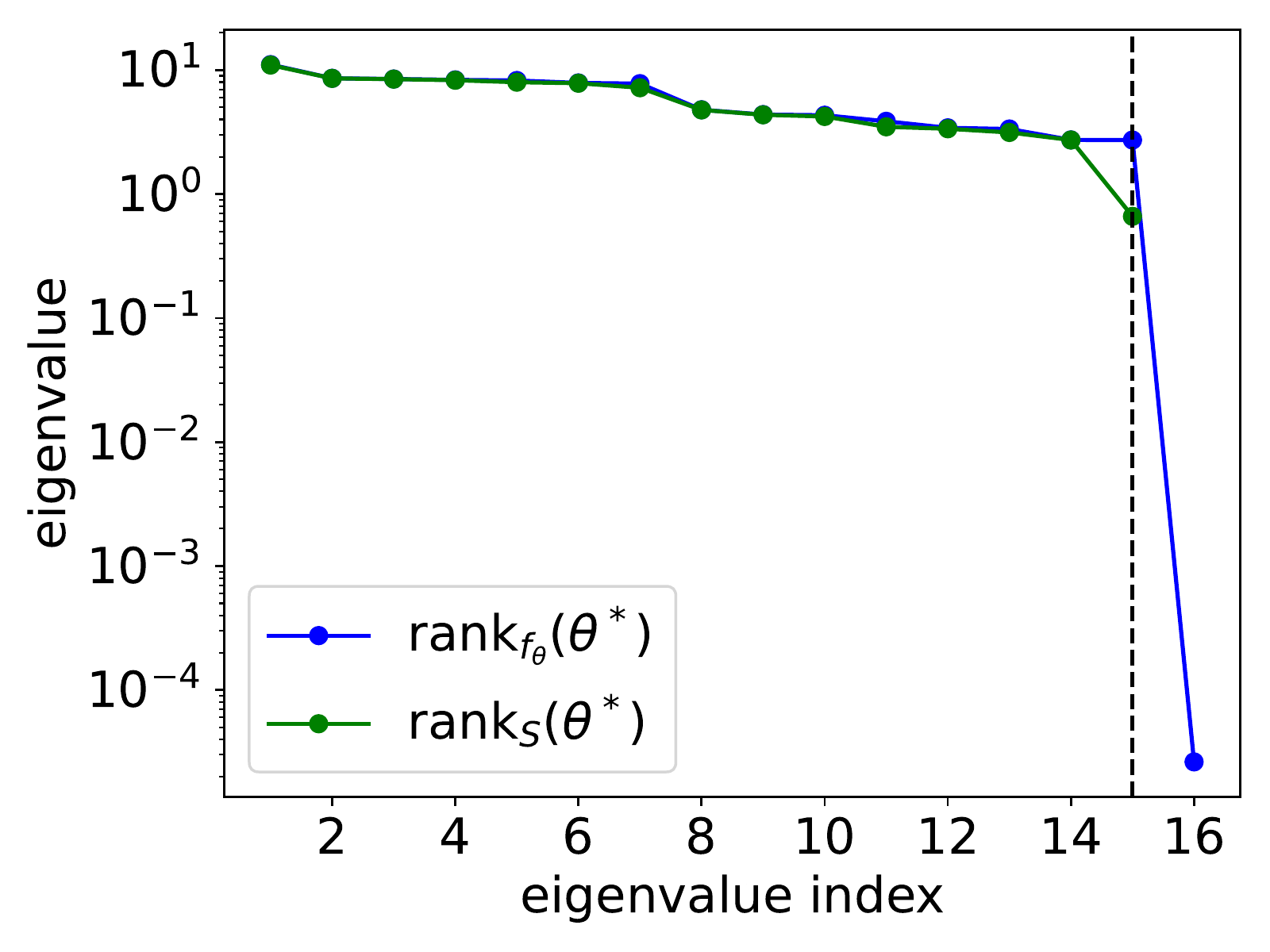}}
  \caption{Learn a $4\times4$ full rank target matrix from different sample sizes. Given training samples of size $n=7, 12, 15$, respectively (indicated by black dashed line in (b-d)), (a)
  average of the ordered eigenvalues of the recovered matrices are presented; (b,c,d) average of the ordered singular values of the empirical tangent matrix $[\partial_{\theta_s}(\vf_{\vtheta^*})_{i_k j_k}]_{s\in[32],k\in[n]}$ vs. that of the tangent matrix $[\partial_{\theta_s}(\vf_{\vtheta^*})_{i_k j_k}]_{s\in[32],k\in[16]}$ ($\{(i_k,j_k)\}_{k=1}^{16}$ takes all of the matrix indices) are presented. Here, $\vtheta^*$ is the recovered parameter vector at convergence. 
  For all experiments, the parameters are initialized by a normal distribution with mean $0$ and variance $10^{-8}$, and trained by full-batch gradient descent with a learning rate $0.05$. Each ordered eigenvalue in the figure is averaged over $50$ ordered eigenvalues obtained from $50$ trials of training with random initialization.} \label{fig:eig_dis}
\end{figure}

\begin{thm}[phase transition of linear stability for recovery, see Theorem \ref{appthm:phase-transition} in Appendix for proof]\label{thm:phase-transition}
Given any analytic model $f_{\vtheta}$, for any target function $f^*\in\fF_{f_{\vtheta}}$ and $n$ generic training data $S=\{(\vx_i,f^*(\vx_i))\}_{i=1}^n$, \\
(i) \textbf{Strictly under-determined regime:} if $n<\rrank_{f_{\vtheta}}(f^*)$, then $f^*$ is not linearly stable;\\
(ii) \textbf{Quasi-determined regime:} if $n\geqslant\rrank_{f_{\vtheta}}(f^*)$, then $f^*$ is linearly stable almost everywhere with respect to $S$.\\
\end{thm}

Above theorem proves a phase transition at $n=\rrank_{f_{\vtheta}}(f^*)$ in general for the linear stability of the target function $f^*$, which coincides with the experimentally observed phase transition in Figs. \ref{fig:random_feature_bar} and \ref{fig:matrix_com}. By this theorem, we can conveniently refer to the fitting problem with training data sizes from $\rrank_{f_{\vtheta}}(f^*)$ to the parameter size $M$ as the quasi-determined regime by default in our paper. 

In general, the linear stability hypothesis indicates an implicit bias of nonlinear training towards linearly stable interpolations, which do not necessarily coincide with the target function. By Lemma \ref{thm:LSC}, any interpolation with model rank higher than data size $n$ is not linearly stable. Thus, we obtain the following corollary showing the implicit bias towards interpolations with lower model ranks by the linear stability hypothesis.
\begin{cor}[implicit bias of linear stability hypothesis, see Corollary \ref{appcor:implicitbias} in Appendix for proof]\label{cor:implicitbias}
Given any model $f_{\vtheta}$ and training data $S=\{(\vx_i,y_i)\}_{i=1}^n$, if an interpolation $f'\in\fF_{f_{\vtheta}}$ is linearly stable, then $\rrank_{f_{\vtheta}}(f')\leqslant n$.
\end{cor}
This bias towards interpolations with lower model ranks highlights the importance of the rank stratification in understanding the fitting behavior of a nonlinear model. When a rank hierarchy is obtained, we immediately obtain a quantitative understanding about the intrinsic preference of the nonlinear model. For example, the rank hierarchy in Table \ref{table:rank_MF} shows that a target with a lower matrix rank has a lower model rank. Therefore the matrix factorization model is intrinsically biased towards a low matrix rank completion by our linear stability hypothesis. We can predict that, even in a strictly under-determined regime, a completion of the matrix with model rank no higher than the sample size is likely to be learned through nonlinear training. In the experiment shown in Fig. \ref{fig:eig_dis}, we find that completions of matrix
 rank $1$, $2$ and $3$ can be reliably learned from $7$, $12$ and $15$ samples, respectively, given a full rank target matrix of size $4\time 4$. In Fig. \ref{fig:eig_dis}(b-d), we further observe that, at these learned parameter points, the empirical model rank equals the model rank. Therefore, despite the failure of recovering the full rank target matrix in these experiments, our linear stability hypothesis and its implicit bias Corollary  \ref{cor:implicitbias} successfully predict the training behavior of the matrix factorization model.

\section{Rank stratification for deep neural networks}\label{sec:DNNs}

By admitting the linear stability hypothesis and establishing the linear stability theory above, the rank hierarchy obtained by rank stratification becomes the key to understand the target recovery performance of a nonlinear model. In this section, we present our rank stratification results for DNNs with numerical demonstration of our theoretical predictions. Our results provide quantitative understandings to the following long-standing open problems: (i) the capability of target recovery for DNNs at overparameterization; (ii) the specialty or superiority of the DNN model in comparison to other models. (iii) the impact of architecture on the target recovery performance of DNNs.
In the following, we first present the rank hierarchies for two-layer tanh-NNs with fully-connected or convolutional architectures and their numerical studies. Later, we provide a rank upper bound  estimate with a partial rank hierarchy for general deep NNs. In the main text, we directly present the results of rank stratification and focus on their implications. All the theoretical details including the rank stratification, supporting theorems and proofs can be found in Appendix Section \ref{appsec:DNN}.

\subsection{Rank stratification for two-layer fully-connected NNs}

In Table~\ref{table:FFN}, we present the rank hierarchy for a two-layer fully-connected tanh-NN with $m$ hidden neurons (see Appendix Section \ref{appsec:FCNN_estimate} for details). Note that similar rank hierarchies can be obtained for two-layer NNs with other architectures and other common activation functions. From Table~\ref{table:FFN}, it is clear that two-layer fully-connected tanh-NNs are rank-adaptive, i.e., different functions occupy different model rank levels, indicating that they are capable of recovering certain target functions at overparameterization. In addition, they possess the following special property---the model expressiveness can be (almost) arbitrarily increased without changing the existing rank hierarchy. For example, when the width of hidden layer increases from $m$ to $m'>m$, the model function space is expanded while the function sets at all model rank levels no greater than $m(d+1)$ remaining unchanged. This is a profound property for a nonlinear model in the following sense. For conventional models, the improvement of expressiveness is in general at a cost of damaging the fitting performance over the original model function space. In linear regression or random feature models, adding a new independent variable or basis function improves the model expressiveness. However, one more data point is needed to recover all functions in the original model function space. Therefore, we always have a hard time to trade off between the model expressiveness and the data size needed for target recovery. However, for the two-layer tanh-NNs, the model rank as an effective size of parameters of any function in the model function space never grows no matter the increase of width $m$. We name this property the \textit{free expressiveness property}. Remark that, this good property is also possessed by NNs with linear or polynomial activations, but their expressiveness cannot be improved to the extent of universal approximation. The practical implication of our result is that, when a two-layer tanh-NN is used for fitting, we do not need to trade the model expressiveness for a good fitting performance. One can simply use a wide NN with sufficient expressiveness even to fit relatively simple target functions without worrying about the generalization performance. 

\begin{table}[htb]
\centering
\renewcommand{\arraystretch}{2.} 
\begin{tabular}{|c|c|}
\hline
\multicolumn{2}{|c|}{model: $f_{\boldsymbol{\theta}}(\boldsymbol{x}) = \sum_{i=1}^{m}a_i\tanh(\boldsymbol{w}_i^\TT \boldsymbol{x}), \boldsymbol{x}\in \mathbb{R}^d, \boldsymbol{\theta} = (a_i, \boldsymbol{w}_i)_{i=1}^m$}                                         \\ \hline
$\rrank_{f_{\vtheta}}(f^*)$                   & \multicolumn{1}{c|}{$f^*$}                            \\ \hline
$0$                                           & \multicolumn{1}{c|}{$0$}                             \\ \hline
$d+1$                                        & \multicolumn{1}{c|}{$\mathcal{F}^{\mathrm{NN}}_1\backslash \{0\}:\{ a_1^*\sigma(\boldsymbol{w}_1^{*\TT}\boldsymbol{x})|a_1^*\neq0,\vw^*_1\neq\vzero\}$}  \\ \hline
$\vdots$                                      & \multicolumn{1}{c|}{$\vdots$}                                                          \\ \hline
$k(d+1)$                                    & \multicolumn{1}{c|}{\makecell[c]{$\mathcal{F}^{\mathrm{NN}}_k\backslash \mathcal{F}^{\mathrm{NN}}_{k-1}$: $\{\sum_{i=1}^k a^*_i\sigma(\boldsymbol{w}_i^{*\TT}\boldsymbol{x})| a_i^*\neq0,\vw^*_i\neq\vzero,$\\ $\vw^*_i\neq\pm\vw^*_j$ for any $i\neq j\}$}} \\ \hline
$\vdots$                                      & \multicolumn{1}{c|}{$\vdots$}                                                     \\ \hline
$m(d+1)$                                        & \multicolumn{1}{c|}{\makecell[c]{$\mathcal{F}^{\mathrm{NN}}_m\backslash \mathcal{F}^{\mathrm{NN}}_{m-1}$: $\{\sum_{i=1}^m a^*_i\sigma(\boldsymbol{w}_i^{*\TT}\boldsymbol{x})| a_i^*\neq0,\vw^*_i\neq\vzero,$\\ $\vw^*_i\neq\pm\vw^*_j$ for any $i\neq j\}$}} \\ \hline
\end{tabular}
\vspace{5pt}
\caption{The rank hierarchy for two-layer fully-connected width-$m$ tanh-NN. \label{table:FFN}}
\end{table}

\subsection{Rank stratification for two-layer CNNs}

In Table \ref{table:model_comparing_simp}, a rank hierarchy is presented in the first two columns for the following simple two-layer tanh-CNN with weight sharing
\begin{align}\label{eq:tanh-CNN_simp}
    f_{\vtheta} (\vx) = \sum_{l=1}^m \sum_{i=1}^{d+1-s} a_{li} \tanh\left(\sum_{\alpha=1}^{s} x_{i+s-\alpha} K_{l;\alpha}\right),\quad  \vx=
  [x_1,\cdots,x_d]^\TT
 \in \sR^{d}.
\end{align}
In addition, to uncover the impact of model architecture on the rank hierarchy, we also estimate the model rank of functions in each function set listed in the first column for the corresponding CNN without weight sharing as well as the corresponding fully-connected NN. All the theoretical details as well as the results for CNNs with 2D image inputs can be found in Appendix Section \ref{appsec:CNN_estimate} and \ref{appsec:arc_comp}. To illustrate the result in Table \ref{table:model_comparing_simp}, we consider a target function generated by a two-layer width-$k$ tanh-CNN with kernel size $3\times3$ and stride $1$ on the MNIST dataset. Without loss of generality, we consider the case without ineffective neurons (whose output weight $a_{li}=0$), i.e., $m_{\mathrm{null}}=0$. Given any $m\geqslant k$, the model rank of this target function is $685k$ in an $m$-kernel CNN, $6760k$ in an $m$-kernel CNN without weight sharing, and $530660k$ in a width-$26m$ fully-connected NN. It is clear that all these NNs possess the free expressiveness property, i.e., the model rank of this target function does not depend on $m$. However, comparing different architectures, the model rank of this target function in a CNN is almost three orders of magnitude less than that in a fully-connected NN. By our linear stability hypothesis and theory, this target function requires $\sim10^3$ times more training data to be recovered by a fully-connect NN than by a CNN. Clearly, regarding the recovery of this target function, the CNN architecture is superior to the CNN architecture without weight sharing, and is far superior to the fully-connect architecture. Note that, the major gap of model rank is between the CNN without weight sharing ($6760k$) and the fully-connect NN ($530660k$). This two orders of magnitude gap of model rank highlights the importance of removing all the unnecessary connections in the design of an NN architecture.

\begin{table}[htb]
\centering
\renewcommand{\arraystretch}{1.5} 
\begin{tabular}{|c|c|c|c|}
\hline
$f^*$                                                           & CNN & CNN without weight sharing                    & Fully-connected NN                            \\ \hline
$0$                                                             & $0$                     & $0$                                           & $0$                                           \\ \hline
$\mathcal{F}_1^{\mathrm{CNN}}\backslash\{0\}$                   & $d+1$       & $(s+1)(d+1-s)-s m_{\mathrm{null}}$  & $(d+1)(d+1-s)-d m_{\mathrm{null}}$  \\ \hline
$\vdots$                                                        & $\vdots$                & $\vdots$                                      & $\vdots$                                      \\ \hline
$\mathcal{F}_k^{\mathrm{CNN}}\backslash\mathcal{F}_{k-1}^{\mathrm{CNN}}$ & $k(d+1)$    & $k(s+1)(d+1-s)-s m_{\mathrm{null}}$ & $k(d+1)(d+1-s)-d m_{\mathrm{null}}$ \\ \hline
$\vdots$                                                        & $\vdots$                & $\vdots$                                      & $\vdots$                                      \\ \hline
$\mathcal{F}_m^{\mathrm{CNN}}\backslash\mathcal{F}_{m-1}^{\mathrm{CNN}}$ & $m(d+1)$    & $m(s+1)(d+1-s)-s m_{\mathrm{null}}$ & $m(d+1)(d+1-s)-d m_{\mathrm{null}}$ \\ \hline
\end{tabular}
\vspace{5pt}
\caption{The rank hierarchy for two-layer tanh-CNN with weight sharing in Eq. \eqref{eq:tanh-CNN_simp}. For functions in each function set over the rank hierarchy, we also present their model rank in the corresponding CNN without weight sharing and the corresponding fully-connected NN. Here $m_{\mathrm{null}}=|\{a_{li}|a_{li}=0\}|$ is a variable counting the number of ineffective neurons in the target function. Note that, when a bias term is added for each hidden neuron (shared or not according to the architecture), the model rank of a function in $\mathcal{F}_k^{\mathrm{CNN}}\backslash\mathcal{F}_{k-1}^{\mathrm{CNN}}$ is $k(d+2)$, $k(s+2)(d+1-s)-s m_{\mathrm{null}}$ and $k(d+2)(d+1-s)-s m_{\mathrm{null}}$, respectively, for these three architectures.} 
\label{table:model_comparing_simp}
\end{table}

\subsection{Experimental demonstration of the linear stability hypothesis in two-layer NNs}
\begin{figure}[htbp]
	\centering
 	\subfigure[different network types]{\includegraphics[width=0.4\textwidth]{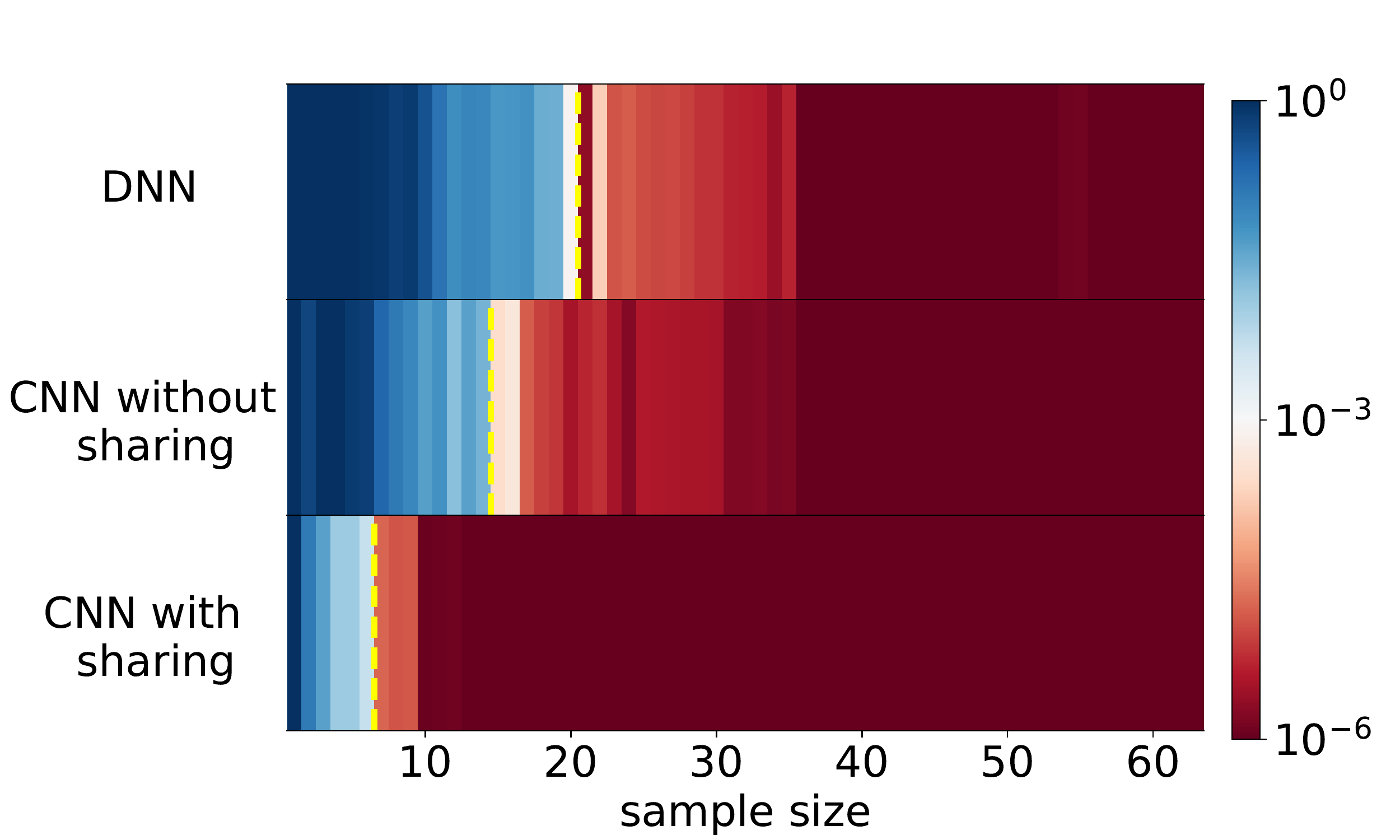}}
	\subfigure[DNN]{\includegraphics[width=0.4\textwidth]{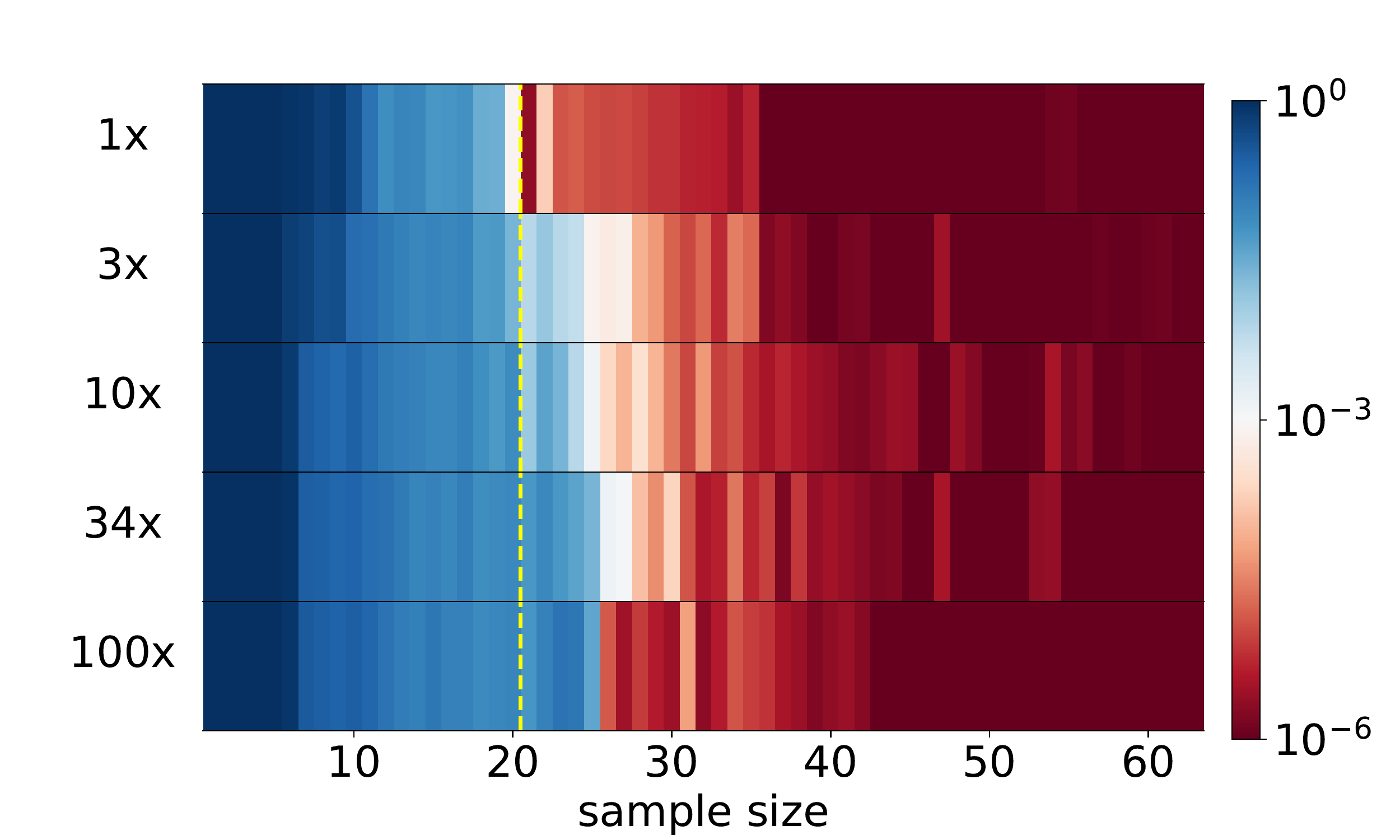}}\\
        \subfigure[CNN without weight sharing]
    {\includegraphics[width=0.4\textwidth]{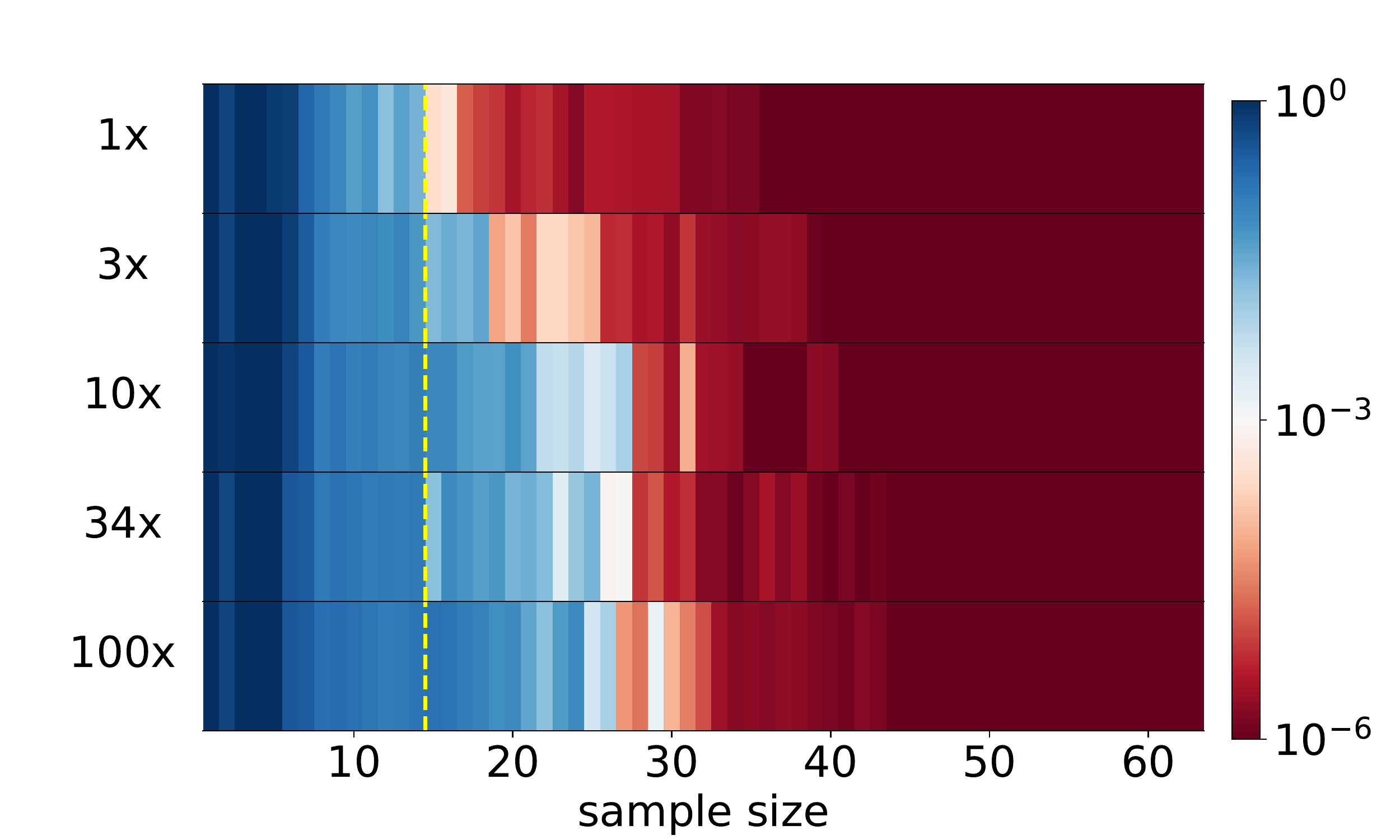}}
        \subfigure[CNN with weight sharing]
    {\includegraphics[width=0.4\textwidth]{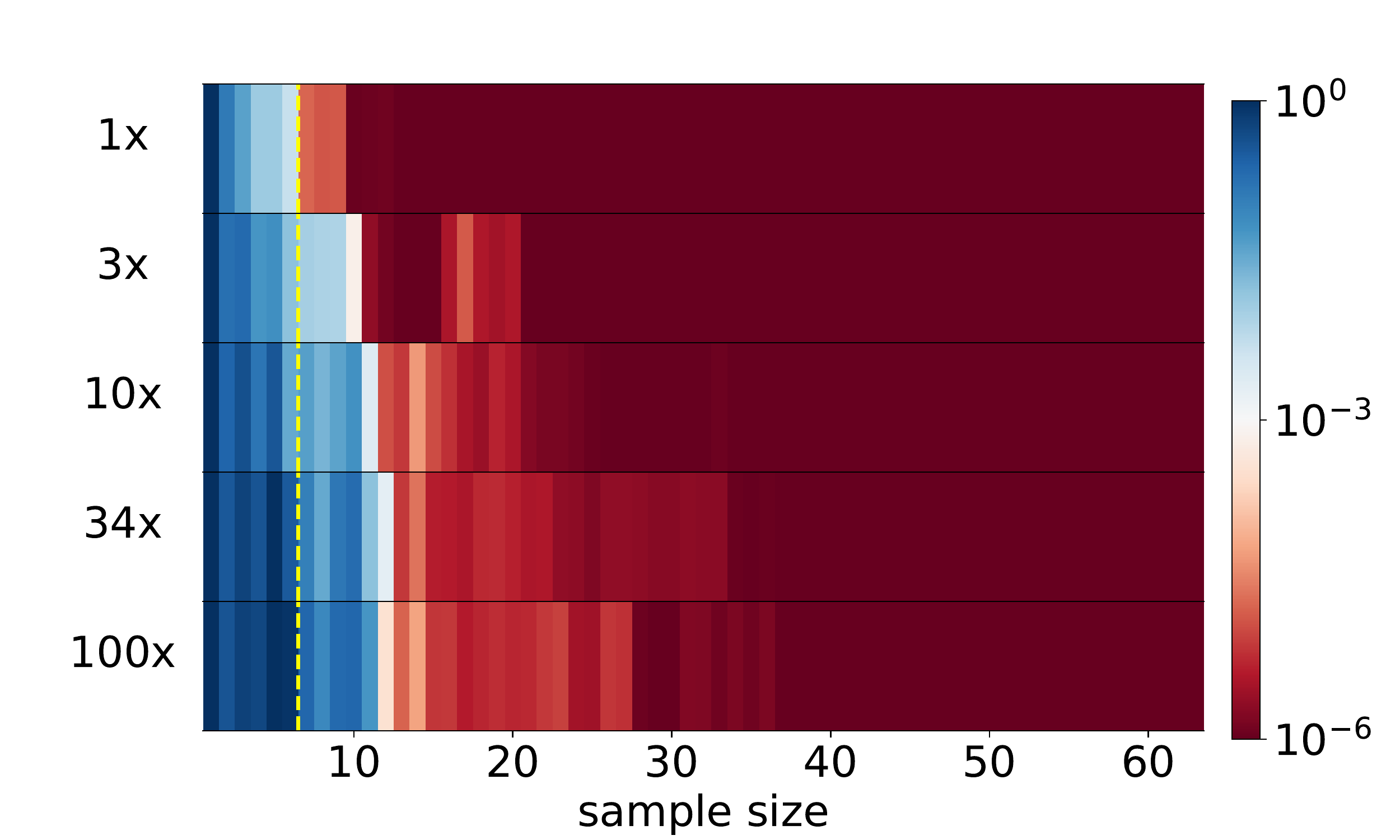}}

\caption{Average test error (color) for NNs of different architectures (ordinate)  and sample sizes (abscissa) in fitting the target function Eq. \eqref{eq:NN_target}. The yellow dashed line for each row indicates the model rank of the target in the corresponding NN. (a) Two-layer $1$-kernel tanh-CNN vs. two-layer $1$-kernel tanh-CNN without weight sharing vs. two-layer width-$3$ fully-connected tanh-NN. Note that these NNs are referred to as $1\textnormal{x}$ for each architecture in (b-d). (b) Two-layer $N$-kernel tanh-CNN, (c) two-layer $N$-kernel tanh-CNN without weight sharing, and (d) two-layer width-$3N$ fully-connected tanh-NN labeled by $N\textnormal{x}$ for $N=1,3,10,34,100$.
For all experiments, network parameters are initialized by a normal distribution with mean $0$ and variance $10^{-20}$, and trained by full-batch gradient descent with a fine-tuned learning rate.} 
\label{fig: network_stab}
\end{figure}

In Fig. \ref{fig: network_stab}, we perform experiments to examine our linear stability hypothesis in two-layer tanh-NNs with different architectures. Specifically, we consider the following target function in our experiments:
\begin{equation}\label{eq:NN_target}
    f^{*}(\boldsymbol{x}) = \boldsymbol{W}^{*[2]}\tanh(\boldsymbol{W}^{*[1]}\boldsymbol{x}),
\end{equation}
where $\boldsymbol{W}^{*[2]}=[1,1,1]$,
$$\boldsymbol{W}^{*[1]}=\left[\begin{array}{cccccc}
0.6 &0.8 &1 & 0 &0 \\
0 &0.6 &0.8 &1 & 0 \\
0 &0 &0.6 &0.8 &1 \\
\end{array}\right]. $$
For the training dataset and the test dataset, we sample the input data from a standard normal distribution and obtain the output values through the target function. We use two-layer tanh-NNs (with a bias term for each hidden neuron) of various architectures and various kernels/widths to fit randomly sampled training datasets of various sizes from $1$ to $63$.
Note that, in a $1$-kernel CNN with or without weight sharing or a width-$3$ fully-connected NN (labeled as $1\textnormal{x}$ in Fig. \ref{fig: network_stab}(b-d)), the model rank of the target function equals the size of model parameters. 
In this situation, the CNN enables a significantly earlier transition of the target recovery accuracy than the other architectures as shown in Fig. \ref{fig: network_stab}(a). However, This result is trivial in the sense that all the recoveries happen at the conventional over-determined/underparameterized regime. 
In Fig. \ref{fig: network_stab}(b-d), we increase the kernels/widths of NNs by $N$ times labeled by $N\textnormal{x}$ for each architecture. Specifically for $N=100$, the sizes of model parameters become $700$, $1500$ and $2100$ respectively. The rank hierarchy in Table \ref{table:model_comparing_simp} gives rise to a constant model rank of $7$, $15$ and $21$ (indicated by yellow dashed lines), respectively, regardless of the choice of $N$. In Fig. \ref{fig: network_stab}(b-d), we observe delayed transitions of the target recovery accuracy for $N>1$, i.e., the test error drops to almost $0$ at a sample size later than the model rank. However, it is easy to notice that the observed transition is far closer to the model rank than to the size of model parameters especially when $N$ is large. We remark that various factors could contribute to a delayed transition of recovery in practice such as a suboptimal tuning of hyperparameters. In practice, it remains an important open problem to find an optimal training method and hyperparameters for NNs to enable a recovery of a target function as close as possible to its model rank.

\subsection{Rank upper bound  estimate for general deep NNs}
For a general DNN model, its rank stratification is difficult. The model rank estimate of any given target function $f^*$ in a DNN function space requires solving the following two challenging problems: (i) identifying the target stratifold $\fM_{f^*}$ in the parameter space; (ii) finding the minimal model rank over the target stratifold. With the help of the previously proposed critical embedding operators \cite{zhang2021embedding,zhang2022embedding}, we can obtain a partial target stratifold and a rank upper bound  estimate for general deep NNs shown in the following theorem (See Appendix Section \ref{appsec:ub_estimate} for details). 

\begin{thm}[rank upper bound  estimate for DNNs, see Theorem \ref{lem:ub_rank} in Appendix for proof]\label{thm:upper_rank_DNN}
Given any NN with $M_{\rwide}$ parameters, for any function $f^*\in\fF_{{\vtheta_{\rnarr}}}$ in the function space of a narrower NN with $M_{\rnarr}$ parameters, we have $\rrank_{f_{\vtheta_{\rwide}}}(f^*)\leqslant \rrank_{f_{\vtheta_{\rnarr}}}(f^*)\leqslant M_{\rnarr}$.
\end{thm}
Here narrower means no larger width in each hidden layer. Applying this theorem to a depth-$L$ width-$\{m_i\}_{i=0}^{L}$ DNN, we obtain a partial rank hierarchy illustrated in Table \ref{tab:partial_rank}. This partial rank hierarchy clearly shows that DNNs are rank-adaptive in general. Specifically, even in a very large DNN with $M_{\rwide}$ parameters, there always exist families of functions with model ranks far less than $M_{\rwide}$, indicating the target recovery capability at heavy overparameterization. Importantly, Theorem \ref{lem:ub_rank} extends the free expressiveness property observed above for two-layer tanh-NNs to general DNNs. By our linear stability hypothesis and theory, this result indicates that one can simply use a wide NN with sufficient expressiveness without worrying about significant deterioration of the target recovery performance.

\begin{table}[htb]
\centering
\renewcommand{\arraystretch}{1.8} 
\begin{tabular}{|c|c|}
\hline
\multicolumn{2}{|c|}{
model:                                                                  $f_{\boldsymbol{\theta}}(\boldsymbol{x}) = \boldsymbol{W}^{[L]}\sigma(\cdots\sigma(\boldsymbol{W}^{[1]}\boldsymbol{x})\cdots)$, $\mW^{[l]}=\sR^{m_l\times m_{l-1}}$,$m_L=1$, $m_0=d$}   \\ \hline
$f^*$                                                                            & upper bound of model rank                                                                                                                                                     \\ \hline
$\mathcal{F}_{\{1, 1, \cdots, 1\}}$                                                           & $d+L-1$                                                                                                                                                                   \\ \hline
$\vdots$                                                                                      & $\vdots$                                                                                                                                                                  \\ \hline
$\mathcal{F}_{\{m_i'\}_{i=1}^{L-1}}, 1\leq m_i' \leq m_i$                                   & $dm_1'+m_1'm_2'+\cdots+m_{L-2}'m_{L-1}' + m_{L-1}'$                                                                                                                 \\ \hline
$\vdots$                                                                                      & $\vdots$                                                                                                                                                                  \\ \hline
$\mathcal{F}_{\{m_i\}_{i=1}^{L-1}}$                                                           & $dm_1+m_1m_2+\cdots+m_{L-2}m_{L-1} + m_{L-1}$                                                                                                                             \\ \hline
\end{tabular}
\vspace{5pt}
\caption{Partial rank hierarchy for general deep fully-connected NNs. $\fF_{\{m_i\}_{i=1}^{L-1}}$ denotes the function space of an $L$-layer DNN with width-$\{m_i\}_{i=1}^{L-1}$ for hidden layers. For simplicity, we consider DNNs without bias terms. \label{tab:partial_rank}}
\end{table}

\section{Conclusions and discussion}\label{sec:conclusion}

\begin{figure}
    \centering
    \includegraphics[width=0.7\textwidth]{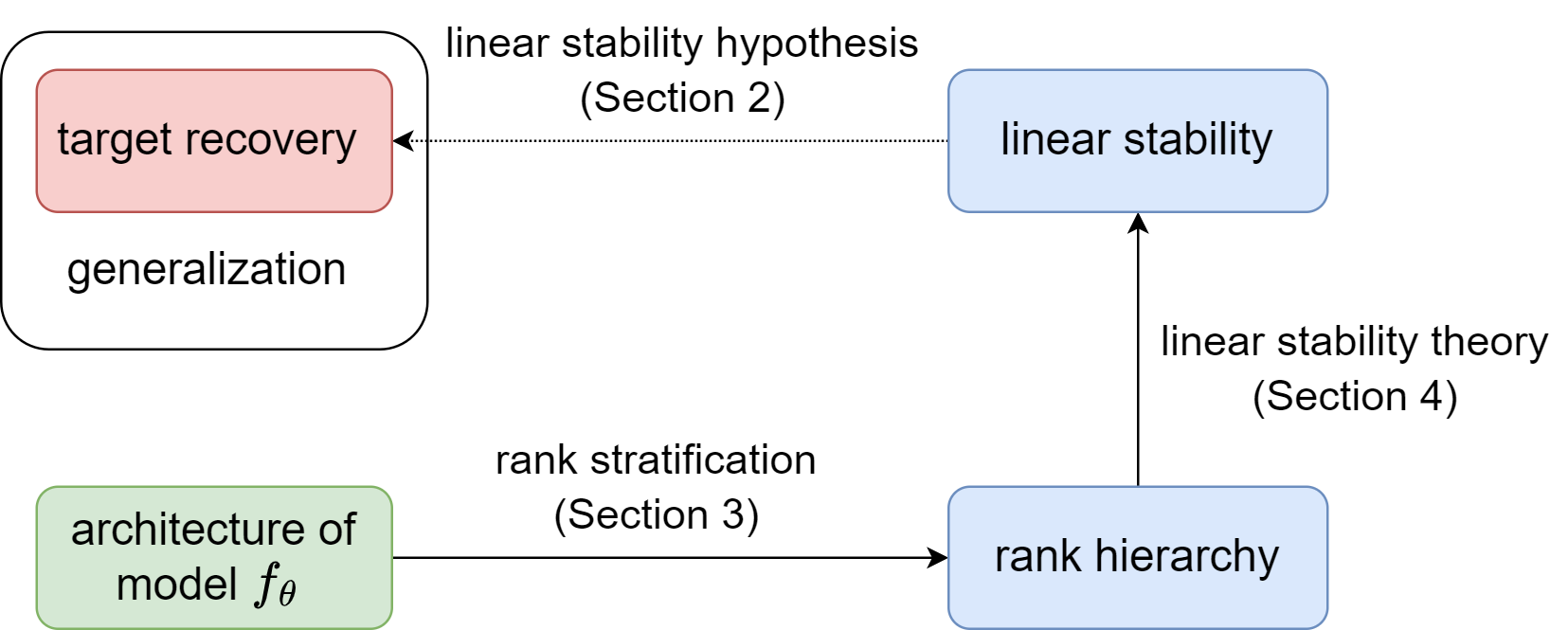}
    \caption{Illustration of the theoretical framework established in our work linking the architecture of a nonlinear model to its quantitative target recovery behavior at overparameterization.}
    \label{fig:flow}
\end{figure}

In this work, we establish a framework to analyze quantitatively the mysterious target recovery behavior at overparameterization for general nonlinear models as illustrated in Fig. \ref{fig:flow}. We apply this framework to the matrix factorization model, two-layer tanh-NNs with a fully-connected or convolutional architecture, and successfully predict their target recovery behaviors even at heavy overparameterization. Remark that our framework relies on a linear stability hypothesis, which needs to be further verified. If this hypothesis is later systematically verified in experiments or even get proved theoretically in certain sense, the following five long standing open problems can be answered quantitatively as follows. Three problems are for general nonlinear models and two problems are specifically for DNNs.\\
\textit{(1) The cause of the target recovery at overparameterization for certain nonlinear models:} a rank-adaptive architecture/parameterization for the nonlinear model. \\
\textit{(2) The effective size of parameters for a nonlinear model:} the model rank quantifies the effective size of parameters for each function in the model function space. Remark that this problem had been proposed by Leo Breiman specifically for NNs almost three decades ago \cite{breiman1995reflections}.\\
\textit{(3) The implicit bias of a nonlinear model through nonlinear training:} towards lower model rank interpolations.\\
\textit{(4) The advantage of the general layer-based architecture of neural networks:} free expressiveness, i.e., expressiveness can be arbitrarily improved through widening with almost no deterioration of the fitting performance.\\
\textit{(5) The superiority of CNNs to fully-connected NNs:} functions in the CNN function space in general possess far lower model ranks in CNNs than in the corresponding fully-connected NNs.\\

In certain sense, our theoretical framework reduces all these important problems to the validity of the linear stability hypothesis. Apart from the evidences provided in this work,
the condensation phenomenon \cite{luo2021phase} during the nonlinear training of DNNs provides a rationale empirical evidence to support the linear stability hypothesis. For a two-layer ReLU NN, the condensation happens when input weights of hidden neurons (the input weights of a hidden neuron consist of all the weights from its input layer and its bias term) condense on isolated orientations. The rank of a condensed ReLU network is independent of, and far less than, the number of network parameters. For two-layer infinite width ReLU networks, \cite{luo2021phase} show that the condensation is a common feature of training networks in the nonlinear regime of the phase diagram (small initialization regime) for both synthetic and real datasets. Similar observations are made for three-layer ReLU NNs \cite{zhou2022towards} and networks with different activation functions \cite{zhou2022empirical}. In addition, large learning rate \cite{andriushchenko2022sgd} and dropout \cite{zhang2022implicit} can facilitate the condensation. Several works provides some preliminary theoretical support for different activations in the initial training stage \cite{maennel2018gradient,pellegrini2020analytic,zhou2022towards}. Therefore, the condensation phenomenon suggests that nonlinear training of neural networks prefers low rank minimizers, which is consistent with the linear stability hypothesis. 
In future works, we will look into details of this hypothesis, e.g., its requirement on the training dynamics for different models, through both experimental and theoretical means.

\section*{Acknowledgement}
This work is sponsored by the National Key R\&D Program of China  Grant No. 2019YFA0709503 (Z. X.), the Shanghai Sailing Program, the Natural Science Foundation of Shanghai Grant No. 20ZR1429000  (Z. X.), the National Natural Science Foundation of China Grant No. 62002221 (Z. X.), the National Natural Science Foundation of China Grant No. 12101401 (T. L.), Shanghai Municipal Science and Technology Key Project No. 22JC1401500 (T. L.), the National Natural Science Foundation of China Grant No. 12101402 (Y. Z.), Shanghai Municipal of Science and Technology Project Grant No. 20JC1419500 (Y.Z.), the Lingang Laboratory Grant No.LG-QS-202202-08 (Y.Z.), Shanghai Municipal of Science and Technology Major Project No. 2021SHZDZX0102, and the HPC of School of Mathematical Sciences and the Student Innovation Center at Shanghai Jiao Tong University.

\bibliography{references}

\begin{thebibliography}{10}

\bibitem{shannon1984communication}
Claude~E Shannon.
\newblock Communication in the presence of noise.
\newblock {\em Proceedings of the IEEE}, 72(9):1192--1201, 1984.

\bibitem{vapnik1998adaptive}
Vladimir~Naumovich Vapnik.
\newblock Adaptive and learning systems for signal processing communications,
  and control.
\newblock {\em Statistical learning theory}, 1998.

\bibitem{bartlett2002rademacher}
Peter~L Bartlett and Shahar Mendelson.
\newblock Rademacher and gaussian complexities: Risk bounds and structural
  results.
\newblock {\em Journal of Machine Learning Research}, 3(Nov):463--482, 2002.

\bibitem{breiman1995reflections}
Leo Breiman.
\newblock Reflections after refereeing papers for nips.
\newblock {\em The Mathematics of Generalization}, XX:11--15, 1995.

\bibitem{zhang2016understanding}
Chiyuan Zhang, Samy Bengio, Moritz Hardt, Benjamin Recht, and Oriol Vinyals.
\newblock Understanding deep learning requires rethinking generalization.
\newblock In {\em 5th International Conference on Learning Representations,
  {ICLR} 2017}.

\bibitem{woodworth2020kernel}
Blake Woodworth, Suriya Gunasekar, Jason~D Lee, Edward Moroshko, Pedro
  Savarese, Itay Golan, Daniel Soudry, and Nathan Srebro.
\newblock Kernel and rich regimes in overparametrized models.
\newblock In {\em Conference on Learning Theory}, pages 3635--3673. PMLR, 2020.

\bibitem{gunasekar2017implicit}
Suriya Gunasekar, Blake~E Woodworth, Srinadh Bhojanapalli, Behnam Neyshabur,
  and Nati Srebro.
\newblock Implicit regularization in matrix factorization.
\newblock {\em Advances in Neural Information Processing Systems}, 30, 2017.

\bibitem{li2018algorithmic}
Yuanzhi Li, Tengyu Ma, and Hongyang Zhang.
\newblock Algorithmic regularization in over-parameterized matrix sensing and
  neural networks with quadratic activations.
\newblock In {\em Conference On Learning Theory}, pages 2--47. PMLR, 2018.

\bibitem{arora2019implicit}
Sanjeev Arora, Nadav Cohen, Wei Hu, and Yuping Luo.
\newblock Implicit regularization in deep matrix factorization.
\newblock {\em Advances in Neural Information Processing Systems}, 32, 2019.

\bibitem{saxe2013exact}
Andrew~M Saxe, James~L McClelland, and Surya Ganguli.
\newblock Exact solutions to the nonlinear dynamics of learning in deep linear
  neural networks.
\newblock {\em arXiv preprint arXiv:1312.6120}, 2013.

\bibitem{lampinen2018an}
Andrew~K. Lampinen and Surya Ganguli.
\newblock An analytic theory of generalization dynamics and transfer learning
  in deep linear networks.
\newblock In {\em International Conference on Learning Representations}, 2019.

\bibitem{gunasekar2018implicit}
Suriya Gunasekar, Jason~D Lee, Daniel Soudry, and Nati Srebro.
\newblock Implicit bias of gradient descent on linear convolutional networks.
\newblock {\em Advances in Neural Information Processing Systems}, 31, 2018.

\bibitem{xu_training_2018}
Zhi-Qin~J Xu, Yaoyu Zhang, and Yanyang Xiao.
\newblock Training behavior of deep neural network in frequency domain.
\newblock {\em International Conference on Neural Information Processing},
  pages 264--274, 2019.

\bibitem{xu2019frequency}
Zhi-Qin~John Xu, Yaoyu Zhang, Tao Luo, Yanyang Xiao, and Zheng Ma.
\newblock Frequency principle: Fourier analysis sheds light on deep neural
  networks.
\newblock {\em Communications in Computational Physics}, 28(5):1746--1767,
  2020.

\bibitem{zhang2021linear}
Yaoyu Zhang, Tao Luo, Zheng Ma, and Zhi-Qin~John Xu.
\newblock A linear frequency principle model to understand the absence of
  overfitting in neural networks.
\newblock {\em Chinese Physics Letters}, 38(3):038701, 2021.

\bibitem{wu2018sgd}
Lei Wu, Chao Ma, and Weinan E.
\newblock How sgd selects the global minima in over-parameterized learning: A
  dynamical stability perspective.
\newblock {\em Advances in Neural Information Processing Systems}, 31, 2018.

\bibitem{mulayoff2021implicit}
Rotem Mulayoff, Tomer Michaeli, and Daniel Soudry.
\newblock The implicit bias of minima stability: A view from function space.
\newblock {\em Advances in Neural Information Processing Systems},
  34:17749--17761, 2021.

\bibitem{zhang2021embedding}
Yaoyu Zhang, Zhongwang Zhang, Tao Luo, and Zhi-Qin~John Xu.
\newblock Embedding principle of loss landscape of deep neural networks.
\newblock {\em Proceedings of the 32nd International Conference on Neural
  Information Processing Systems}, 2021.

\bibitem{zhang2022embedding}
Yaoyu Zhang, Yuqing Li, Zhongwang Zhang, Tao Luo, and Zhi-Qin~John Xu.
\newblock Embedding principle: a hierarchical structure of loss landscape of
  deep neural networks.
\newblock {\em Journal of Machine Learning vol}, 1:1--45, 2022.

\bibitem{luo2021phase}
Tao Luo, Zhi-Qin~John Xu, Zheng Ma, and Yaoyu Zhang.
\newblock Phase diagram for two-layer relu neural networks at infinite-width
  limit.
\newblock {\em J. Mach. Learn. Res.}, 22:71--1, 2021.

\bibitem{zhou2022towards}
Hanxu Zhou, Qixuan Zhou, Tao Luo, Yaoyu Zhang, and Zhi-Qin~John Xu.
\newblock Towards understanding the condensation of neural networks at initial
  training.
\newblock {\em Advances in Neural Information Processing Systems}, 2022.

\bibitem{zhou2022empirical}
Hanxu Zhou, Qixuan Zhou, Zhenyuan Jin, Tao Luo, Yaoyu Zhang, and Zhi-Qin~John
  Xu.
\newblock Empirical phase diagram for three-layer neural networks with infinite
  width.
\newblock {\em Advances in Neural Information Processing Systems}, 2022.

\bibitem{andriushchenko2022sgd}
Maksym Andriushchenko, Aditya Varre, Loucas Pillaud-Vivien, and Nicolas
  Flammarion.
\newblock Sgd with large step sizes learns sparse features.
\newblock {\em arXiv preprint arXiv:2210.05337}, 2022.

\bibitem{zhang2022implicit}
Zhongwang Zhang and Zhi-Qin~John Xu.
\newblock Implicit regularization of dropout.
\newblock {\em arXiv preprint arXiv:2207.05952}, 2022.

\bibitem{maennel2018gradient}
Hartmut Maennel, Olivier Bousquet, and Sylvain Gelly.
\newblock Gradient descent quantizes relu network features.
\newblock {\em arXiv preprint arXiv:1803.08367}, 2018.

\bibitem{pellegrini2020analytic}
Franco Pellegrini and Giulio Biroli.
\newblock An analytic theory of shallow networks dynamics for hinge loss
  classification.
\newblock {\em Advances in Neural Information Processing Systems}, 33, 2020.

\bibitem{fukumizu2019semi}
Kenji Fukumizu, Shoichiro Yamaguchi, Yoh-ichi Mototake, and Mirai Tanaka.
\newblock Semi-flat minima and saddle points by embedding neural networks to
  overparameterization.
\newblock {\em Advances in Neural Information Processing Systems},
  32:13868--13876, 2019.

\bibitem{csimcsek2021geometry}
Berfin Simsek, Fran{\c{c}}ois Ged, Arthur Jacot, Francesco Spadaro, Clement
  Hongler, Wulfram Gerstner, and Johanni Brea.
\newblock Geometry of the loss landscape in overparameterized neural networks:
  Symmetries and invariances.
\newblock In {\em Proceedings of the 38th International Conference on Machine
  Learning}, pages 9722--9732. PMLR, 2021.

\bibitem{bai2022embedding}
Zhiwei Bai, Tao Luo, Zhi-Qin~John Xu, and Yaoyu Zhang.
\newblock Embedding principle in depth for the loss landscape analysis of deep
  neural networks.
\newblock {\em arXiv preprint arXiv:2205.13283}, 2022.

\end{thebibliography}
\bibliographystyle{unsrt}

\appendix
\section{Details of rank stratification}

\begin{definition}[model rank]
Given any differentiable (in parameters) model $f_{\vtheta}$, the model rank for any $\vtheta^*\in \sR^M$ is defined as
\begin{equation}
\rrank_{f_{\vtheta}}(\vtheta^*):=\rdim\left(\rspan\left\{\partial_{\theta_i} f(\cdot;\vtheta^*)\right\}_{i=1}^M\right),
\end{equation}
where $\rspan\left\{ \phi_i(\cdot)\right\}_{i=1}^M=\{\sum_{i=1}^M a_i\phi_i(\cdot)|a_1,\cdots,a_M\in\sR\}$ and $\rdim(\cdot)$ returns the dimension of a linear function space. Then the model rank for any function $f^*\in\fF_{f_{\vtheta}}$ with model function space $\fF_{f_{\vtheta}}:=\{f(\cdot;\vtheta)|\vtheta\in\sR^M\}$ is defined as 
\begin{equation}
    \rrank_{f_{\vtheta}}(f^*):=\min_{\vtheta'\in\fM_{f^*}}\rrank_{f_{\vtheta}}(\vtheta'),
\end{equation}
where the target stratifold $\fM_{f^*}:=\{\vtheta|f(\cdot;\vtheta)=f^*;\vtheta\in\sR^M\}$.
\end{definition}

Given a differentiable model $f_{\vtheta}$, the standard procedure of rank stratification is comprised of the following two steps:\\
\textbf{Step 1:} Stratify the parameter space into different model rank levels to obtain the rank hierarchy over the parameter space;\\
\textbf{Step 2:} Stratify the model function space into different model rank levels to obtain the rank hierarchy over the model function space.\\
The difficulty of rank stratification depends on the complexity of model architecture. For example, this standard two-step rank stratification is straight-forward for $f_{\rNL}=\theta_0+\theta_1 x_1 +\theta_2\theta_3 x_2$ as illustrated in the main text. For a matrix factorization model, a linear algebra lemma is needed for its rank stratification. For DNN models, rank stratification is in general difficult. In our work, with the help of the previously discovered embedding principle and critical embedding operators, we obtain a complete rank hierarchy for two-layer tanh-NNs and a partial rank hierarchy for general multi-layer DNNs.

\subsection{Matrix factorization\label{appsec:MF}}

\subsubsection{Theoretical preparation}
\begin{lemma}[linear algebra lemma]
Let $\mA$ and $\mB$ be two ($d\times d$) matrices and $r_{\mA}:= \rrank(\mA), r_{\mB} := \rrank(B)$,
$$
\mGamma= \left[
 \begin{array}{c}
   \mI \otimes \mB \\
   \mA^{\TT}\otimes  \mI \\
  \end{array}
  \right],
$$
where $\mI$ is the ($d\times d$) identity matrix and $\otimes$ is the Kronecker product.
Then $\rrank(\mGamma) = d^2 - (d-r_{\mA})(d-r_{\mB}).$
\label{lem:matrix_completion}
\end{lemma}
\begin{proof}
In order to compute the rank of $\mGamma$, we consider the dimension of the null space $N(\mGamma)$ of $\mGamma$ due to the relationship 
$$\operatorname{rank}(\mGamma) + \operatorname{dim}(N(\mGamma)) = d^2.$$
We will show that $\operatorname{dim}(N(\mGamma)) = (d-r_{\mA})(d-r_{\mB})$, thus $\operatorname{rank}(\mGamma) = d^2 - (d-r_{\mA})(d-r_{\mB})$, as desired.

Let $n_{\mA} = d -r_{\mA}, n_{\mB} = d -r_{\mB}$, and suppose that $N(\mA^\TT)=\operatorname{span}\{\alpha_1, \alpha_2, \cdots, \alpha_{n_{\mA}}\}$, $N(\mB) = \operatorname{span}\{\beta_1, \beta_2, \cdots, \beta_{n_{\mB}}\}$ are the null spaces of $\mA^\TT$ and $\mB$, respectively. 
Since for any $1\leq i\leq n_{\mA}, 1\leq j \leq n_{\mB}$
$$ \left[
 \begin{matrix}
   \mI \otimes \mB \\
   \mA^{\TT}\otimes  \mI \\
  \end{matrix}
  \right]  \left[
 \begin{matrix}
   \alpha_i\otimes\beta_j
  \end{matrix}
  \right] = \left[
 \begin{matrix}
   \mI\alpha_i \otimes \mB\beta_j \\
   \mA^{\TT}\alpha_i\otimes  \mI\beta_j \\
  \end{matrix}
  \right] = \left[
 \begin{matrix}
   \alpha_i \otimes \boldsymbol{0} \\
   \boldsymbol{0}\otimes  \beta_j \\
  \end{matrix}
  \right]= \left[
 \begin{matrix}
   \boldsymbol{0} \\
   \boldsymbol{0} \\
  \end{matrix}
  \right],$$
we have $N(\mA^\TT)\otimes N(\mB)\subseteq N(\mGamma)$.

On the other hand, let $\boldsymbol{0} \neq \boldsymbol{x} \in \mathbb{R}^{d^2}$ be in the null space $N(\mGamma)$ of $\mGamma$, i.e. $\mGamma \vx = \boldsymbol{0}$. We have
$$ \left[
 \begin{matrix}
   \boldsymbol{0} \\
   \boldsymbol{0} \\
  \end{matrix}
  \right] = \left[
 \begin{matrix}
   \mI \otimes \mB \\
   \mA^{\TT}\otimes  \mI \\
  \end{matrix}
  \right]  \vx
  = \left[
 \begin{matrix}
   (\mI \otimes \mB)\vx \\
   (\mA^{\TT}\otimes  \mI)\vx \\
  \end{matrix}
  \right] = \left[
 \begin{matrix}
   \operatorname{vec}(\mB \mX\mI^\TT) \\
   \operatorname{vec}(\mI\mX \mA) \\
  \end{matrix}
  \right] = \left[
 \begin{matrix}
   \operatorname{vec}(\mB \mX) \\
   \operatorname{vec}(\mX\mA) \\
  \end{matrix}
  \right],$$
where $\mX$ is the inverse of the vectorization operator (formed by reshaping the vector $\vx = [x_1, x_2, \ldots, x_{d^2}]^{\TT}$), namely,
$$
\mX = \left[
\begin{matrix}
  x_1 & x_{d+1} & \cdots & x_{(d-1)d+1}\\
  x_2 & x_{d+2} & \cdots & x_{(d-1)d+2}\\
  \vdots & \vdots & \ddots & \vdots\\
  x_d & x_{d+d} & \cdots & x_{d^2}
\end{matrix}\right].
$$

Therefore, we conclude $\mB\mX = \boldsymbol{0}$ and $\mA^\TT \mX^\TT = \boldsymbol{0}$. Note that the first equation $\mB\mX = \boldsymbol{0}$ implies each column of $\mX$ is a linear combination of $\{\beta_1, \beta_2, \cdots, \beta_{n_{\mB}}\}$. Thus, there exists $\mC_{n_{\mB}\times d}$ such that $$\mX = \left[\beta_1, \beta_2, \cdots, \beta_{n_{\mB}}\right]\mC.$$

Since $\{\beta_1, \beta_2, \cdots, \beta_{n_{\mB}}\}$ is linearly independent, the second equation $\mA^\TT \mX^\TT = \boldsymbol{0}$ implies $\mA^\TT\mC^{\TT} = \vzero.$ Thus, the $i$-th row $\mC_{i}$ of matrix $\mC$ satisfies $\mC_i\in N(\mA^{\TT})$ for any $i\in [n_{\mB}]$. By re-vectorizing $\vx = \mathrm{vec}(\mX) = [x_1, x_2, \ldots, x_{d^2}]^{\TT}$, we have
$$
\vx = \mC_{1}\otimes \beta_1 + \mC_2\otimes \beta_2 + \cdots + \mC_{n_{\mB}}\otimes \beta_{n_{\mB}}.
$$
Therefore, we conclude that $\boldsymbol{x} \in N(\mA^\TT)\otimes N(\mB)$ and $N(\mGamma) \subseteq N(\mA^\TT)\otimes N(\mB)$.

Now we have $N(\mGamma) = N(\mA^\TT)\otimes N(\mB)$, by which
$$ \operatorname{dim}(N(\mGamma)) = \operatorname{dim}(N(\mA)\otimes N(\mB)) = (d-r_{\mA})(d-r_{\mB}), $$
and $\operatorname{rank}(\mGamma) = d^2 - (d-r_{\mA})(d-r_{\mB})$. 
\end{proof}

\subsubsection{Rank stratification}

\textbf{Matrix factorization model:} $\boldsymbol{f}_{\boldsymbol{\theta}} = \mA \mB$ with $\vtheta=(\mA,\mB)$, $\mA,\mB\in\sR^{d\times d}$.

\textbf{Step 1:} Stratify the parameter space into different model rank levels to obtain the rank hierarchy over the parameter space.

Given any parameter point $\mA = [a_{ij}]_{i, j=1}^{d}\in\sR^{d\times d}, \mB = [b_{ij}]_{i, j=1}^{d}\in\sR^{d\times d}$. Consider the tangent space 
$$
\operatorname{span} \big\{\mP^{ij}, \mQ^{ij}\big\}_{i, j=1}^{d},
$$
where $\mP^{ij} = \dfrac{\partial\boldsymbol{f}_{\vtheta}}{\partial a_{ij}}$, $\mQ^{ij} = \dfrac{\partial\boldsymbol{f}_{\vtheta}}{\partial b_{ij}}$ and $\mathrm{rank}(\mP^{ij}) = \mathrm{rank}(\mQ^{ij}) = 1$.

By vectorizing $\mP^{ij}_{d\times d}$ and $\mQ^{ij}_{d\times d}$, we get $$\mathrm{vec}(\mP^{ij}) = [P^{ij}_{11}, \cdots, P^{ij}_{1d}, P^{ij}_{21}, \cdots, P^{ij}_{2d}, \cdots, P^{ij}_{d1}, \cdots, P^{ij}_{dd}]^{\TT}\in \sR^{d^2},$$
$$
\mathrm{vec}(\mQ^{ij}) = [Q^{ij}_{11}, \cdots, Q^{ij}_{1d}, Q^{ij}_{21}, \cdots, Q^{ij}_{2d}, \cdots, Q^{ij}_{d1}, \cdots, Q^{ij}_{dd}]^{\TT}\in \sR^{d^2}.
$$ 
Now we put these vectors into a matrix $\mGamma_{2d^2\times d^2}$, namely
$$ \mGamma_{2d^2\times d^2} = \left[
  \begin{matrix}
  \mathrm{vec}(\mQ^{11}), & \cdots,  \mathrm{vec}(\mQ^{dd}), & \mathrm{vec}(\mP^{11}), & \cdots, & \mathrm{vec}(\mP^{dd})
  \end{matrix}  \right]^{\TT}. 
$$
Clearly, 
$$\operatorname{rank}(\mGamma) = \operatorname{dim}\left(\operatorname{span} \big\{\mP^{ij}, \mQ^{ij}\big\}_{i, j=1}^{d}\right).
$$
Therefore, we only need to compute the rank of matrix $\mGamma$.

By exploiting the Kronecker product of matrices, we are able to write $\mGamma$ in a more concise form:
$$
\mGamma= \left[\begin{array}{cccc}
\mB & & & \\
& \mB & & \\
& & \ddots & \\
& & & \mB \\
a_{11} \mI & a_{21} \mI & \cdots & a_{d 1} \mI \\
a_{12} \mI & a_{22} \mI & \cdots & a_{d 2} \mI \\
\vdots & \vdots & \ddots & \vdots \\
a_{1 d} \mI & a_{2 d} \mI & \cdots& a_{d d} \mI
\end{array}\right] = \left[
 \begin{matrix}
   \mI \otimes \mB \\
   \mA^{\TT}\otimes  \mI \\
  \end{matrix}
  \right].
$$

Let $r_{\mA}:= \rrank(\mA), r_{\mB} := \rrank(\mB)$. By Lemma~\ref{lem:matrix_completion}, the rank of $\mGamma$ is $d^2 - (d-r_{\mA})(d-r_{\mB})$. Therefore the matrix factorization model possesses the rank levels $\{d^2 - (d-r_1)(d-r_2)|r_1,r_2\in[d]\}$ over its parameter space, each of which is occupied by $\{(\mA,\mB)|\rrank(\mA)=r_1,\rrank(\mB)=r_2\}$.

\textbf{Step 2:} Stratify the model function space into different model rank levels to obtain the rank hierarchy over the model function space.

Given any matrix $\vf^*\in\sR^{d\times d}$, let $r=\rrank(\vf^*)$. By definition, the model rank of $\boldsymbol{f}^*$ is the minimal model rank among all parameters recovering $\boldsymbol{f}^*$. 
Because
\begin{equation*}
    \rrank(\mA\mB)\leq \min\{\rrank(\mA), \rrank(\mB)\},
\end{equation*}
any factorization $\vf^*=\mA^*\mB^*$ satisfies $\rrank(\mA^*)\geqslant r$ and $\rrank(\mB^*)\geqslant r$. By the analysis in Step 1, we have $ \rrank_{\vf_{\vtheta}}(\vtheta^*) \geqslant d^2-(d-r)^2=2rd-r^2$. By the singular value decomposition $\vf^*=\mU\mSigma\mV^\TT$, $\mA^*=\mU\mSigma^{\frac{1}{2}}$ and $\mB^*=\mSigma^{\frac{1}{2}}\mV^\TT$ recover $\vf^*$ with $\rrank(\mA^*)=\rrank(\mB^*)=r$. Therefore, $\rrank_{\vf_{\vtheta}}(\vtheta^*)$ attains its lower bound $2rd-r^2$, thus $\rrank_{\vf_{\vtheta}}(\vf^*)=2rd-r^2$. Then, the matrix factorization model possesses the rank levels $\{2rd-r^2|r\in[d]\}$ over its function space, each of which is occupied by $\{\vf\in\sR^{d\times d}|\rrank(\vf)=r\}$ as illustrated in Table \ref{table:app_rank_MF}.

\begin{table}[htb]
\caption{Rank hierarchy for the matrix factorization model.\label{table:app_rank_MF}}
\centering
\renewcommand{\arraystretch}{1.5}
\begin{tabular}{|c|cc|}
\hline
model                                         & \multicolumn{2}{c|}{$\vf_{\vtheta} = \mA\mB, \vtheta = (\mA, \mB), \mA, \mB \in \sR^{d\times d}$}                                         \\ \hline
$\operatorname{rank}_{\vf_{\vtheta}}(\vf^{*})$                  & \multicolumn{1}{c|}{$\vf^*$}                        & $\argmin_{\vtheta'\in\fM_{f^*}}\rrank_{f_{\vtheta}}(\vtheta')$                                         \\ \hline
$0$                                           & \multicolumn{1}{c|}{$\boldsymbol{0}_{d\times d}$}               & $\mA = \mB = \boldsymbol{0}_{d\times d}$                            \\ \hline
$2d-1$                                        & \multicolumn{1}{c|}{$\operatorname{rank}(\vf^*) = 1$} & $\mathrm{rank}(\mA) = \mathrm{rank}(\mB) = 1, \mA\mB=\vf^*$ \\ \hline
$\vdots$                                      & \multicolumn{1}{c|}{$\vdots$}                       & $\vdots$                                            \\ \hline
$2rd-r^2$                                     & \multicolumn{1}{c|}{$\operatorname{rank}(\vf^*) = r$} & $\mathrm{rank}(\mA) = \mathrm{rank}(\mB) = r, \mA\mB=\vf^*$ \\ \hline
$\vdots$                                      & \multicolumn{1}{c|}{$\vdots$}                       & $\vdots$                                            \\ \hline
$d^2$                                         & \multicolumn{1}{c|}{$\operatorname{rank}(\vf^*) = d$} & $\mathrm{rank}(\mA) = \mathrm{rank}(\mB) = d, \mA\mB=\vf^*$ \\ \hline
\end{tabular}
\end{table}
Remark that the above analysis serves as a proof of the following proposition.
\begin{prop}[rank hierarchy of a matrix factorization model] Given a matrix factorization model $\boldsymbol{f}_{\boldsymbol{\theta}} = \mA \mB$ with $\vtheta=(\mA,\mB)$, $\mA,\mB\in\sR^{d\times d}$, 
for any matrix $\vf^*\in\sR^{d\times d}$, we have model rank
$$\rrank_{\vf_{\vtheta}}(\vf^*)=2rd-r^2,$$ where $r=\rrank(\vf^*)$ is the matrix rank of $\vf^*$.
\end{prop}

\subsection{DNNs\label{appsec:DNN}}

\subsubsection{Rank upper bound  estimate via critical mappings\label{appsec:ub_estimate}}
Rank stratification for DNNs is in general difficult. Luckily, the recent discovery of the embedding principle and the critical embedding operators provide powerful tools for the rank stratification \cite{zhang2021embedding,zhang2022embedding,fukumizu2019semi,csimcsek2021geometry,bai2022embedding}. In the following, we first provide a general definition of critical mappings, by which the previously proposed critical embeddings are special cases. Then, we prove Lemma \ref{lem:ub_rank}, showing that uncovering critical mappings is an important means for obtaining an upper bound estimate of the model rank. This general result combined with the embedding principle directly provides a rank upper bound estimate for general deep NNs illustrated in Table. \ref{tab:partial_rank}.

\begin{definition}[critical mapping]
Given two differentiable models $f_{\vtheta_A}=f(\cdot;\vtheta_A)$ with $\vtheta_A\in\sR^{M_A}$ and $g_{\vtheta_B}=g(\cdot;\vtheta_B)$ with $\vtheta_B\in\sR^{M_B}$, $\fP:\sR^{M_A}\to\sR^{M_B}$ is a critical mapping from model $A$ to $B$ if given any $\vtheta\in \sR^{M_A}$, we have\\
(i) output preserving: $f_{\vtheta}=g_{\fP(\vtheta)}$;\\
(ii) criticality preserving: for any data $S=\{(\vx_i,y_i)\}_{i=1}^n$ and empirical risk function $R_S(\cdot)$, if $\nabla_{\vtheta}R_S(f_{\vtheta})=\vzero$, then $\nabla_{\vtheta}R_S(g_{\fP(\vtheta)})=\vzero$.
\end{definition}

\begin{lemma}[rank upper bound  estimate]\label{lem:ub_rank}
Given two models $f_{\vtheta_A}=f(\cdot;\vtheta_A)$ with $\vtheta_A\in\sR^{M_A}$ and $g_{\vtheta_B}=g(\cdot;\vtheta_B)$ with $\vtheta_B\in\sR^{M_B}$, if there exists a critical mapping $\fP$ from model $A$ to $B$, then $\rrank_g(f^*)\leqslant\rrank_f(f^*)\leqslant M_A$ for any $f^*\in\fF_{f}$.
\end{lemma}
\begin{rmk}
If $M_B\gg M_A$, this upper bound estimate is highly informative, indicating target recovery capability at heavy overparameterization for model $B$.  Importantly, this lemma establishes the relation between our rank stratification and previous studies about the critical embedding for the DNN loss landscape analysis. As a result, the critical embedding intrinsic to the DNN architecture not only benefits optimization as studied in previous works, but also profoundly benefits the recovery/generalization performance.
\end{rmk}
\begin{proof}
 By Definition \ref{def:rank_theta}, for any $f^*\in\fF_{f}$, there exists $\vtheta^*\in\sR^{M_A}$ such that $\rrank_f(\vtheta^*)=\rrank_f(f^*)$. Then, $g_{\fP(\vtheta^*)}=f_{\vtheta^*}=f^*$. Without loss of generality, we consider $R_S(h)=\frac{1}{2}\sum_{i=1}^n(h(\vx_i)-y_i)^2$. Then $\nabla_{\vtheta}R_S(f_{\vtheta^*})=\sum_{i=1}^n (y_i-f^*(\vx_i))\nabla_{\vtheta^*}f_{\vtheta^*}(\vx_i)$ and $\nabla_{\fP(\vtheta)}R_S(g_{\fP(\vtheta^*)})=\sum_{i=1}^n (y_i-f^*(\vx_i))\nabla_{\fP(\vtheta)}g_{\fP(\vtheta^*)}(\vx_i)$. Because $\fP$ is criticality preserving for arbitrary data $S$, we have $\ker(\nabla_{\vtheta}f_{\vtheta^*}(\mX))\subseteq\ker(\nabla_{\fP(\vtheta)}g_{\fP(\vtheta^*)}(\mX))$ for any $\mX:=[\vx_1,\cdots,\vx_n]$. Here, $\nabla_{\vtheta}f_{\vtheta^*}(\mX)=[\nabla_{\vtheta}f_{\vtheta^*}(\vx_1),\cdots,\nabla_{\vtheta}f_{\vtheta^*}(\vx_n)]$. Because $\rrank_S(\fP(\vtheta^*))+\rdim(\ker(\nabla_{\fP(\vtheta)}g_{\fP(\vtheta^*)}(\mX)))=\rrank_S(\vtheta^*)+\rdim(\ker(\nabla_{\vtheta}f_{\vtheta^*}))=n$, we have $\rrank_S(\fP(\vtheta^*))\leqslant \rrank_S(\vtheta^*)$ for any data $S$ (for the definition of $\rrank_{S}(\vtheta^*)$, see Appendix Section \ref{appsec:LST}). Taking the infinite data limit, we obtain $\rrank_g(\fP(\vtheta^*))\leqslant\rrank_f(\vtheta^*)\leqslant M_A$. Therefore, $\rrank_g(f^*)\leqslant\rrank_f(f^*)\leqslant M_A$ for any $f^*\in\fF_{f}$.
\end{proof}

\begin{theorem}[Embedding Principle, Theorem 4.2 in Ref.  \cite{zhang2022embedding}]\label{thm:embeddingPrinciple}
Given any NN and any  $K$-neuron wider NN, there exists a $K$-step composition embedding $\fP$ satisfying  that:
For any given data $S$, loss function $\ell(\cdot,\cdot)$, activation function $\sigma(\cdot)$,  given any   critical point $\vtheta^{\rc}_{\rnarr}$ of the narrower NN,   $\vtheta^{\rc}_{\rwide}:=\fP(\vtheta^{\rc}_{\rnarr})$ is still a critical point of the  $K$-neuron wider NN with the same output function, i.e., $\vf_{\vtheta^{\rc}_{\rnarr}}=\vf_{\vtheta^{\rc}_{\rwide}}$.
\end{theorem}
Here wider/narrower means no smaller/larger width in each hidden layer (see either of Refs. \cite{zhang2021embedding,zhang2022embedding} for a mathematical definition). As a direct consequence of Lemma \ref{lem:ub_rank} and Theorem \ref{thm:embeddingPrinciple}, we obtain the following theorem.
\begin{thm}[rank upper bound  estimate for DNNs, Theorem \ref{thm:upper_rank_DNN} in the main context]
Given any NN with $M_{\rwide}$ parameters, for any function in the function space of a narrower NN with $M_{\rnarr}$ parameters $f^*\in\fF_{{\vtheta_{\rnarr}}}$, we have $\rrank_{f_{\vtheta_{\rwide}}}(f^*)\leqslant \rrank_{f_{\vtheta_{\rnarr}}}(f^*)\leqslant M_{\rnarr}$.
\label{thm:app_embedding}
\end{thm}
This theorem gives rise to the partial rank hierarchy exhibited in Table. \ref{tab:partial_rank}.

\subsubsection{Theoretical preparation for two-layer NN rank stratification\label{appsec:two-layer_thm}}

\begin{prop}\label{appendix prop: linear independence of activations}
Let $\sigma: \sR \to \sR$ be any analytic function such that $\sigma^{(n_j)}(0) \neq 0$ for an infinite sequence of distinct indices $\{n_j\}_{j=1}^\infty$. Given $d \in \sN$ and $m$ distinct weights $\vw_1, ..., \vw_m \in \sR^d \backslash \{\vzero\}$, such that $\vw_k\neq\pm \vw_j$ for all $1 \leqslant k < j \leqslant m$. Then $\{\sigma(\vw_i^\TT \vx), \sigma'(\vw_i^\TT \vx)x_1, ..., \sigma'(\vw_i^\TT \vx)x_d\}_{i=1}^m$ is a linearly independent function set.  
\end{prop}
\begin{proof}
For $x$ sufficiently close to $0 \in \sR$, we can write $\sigma(x) = \sum_{j=0}^\infty c_j x^j$, where $c_j = \sigma^{(j)}(0)/(j!)$. Then, $\sigma'(x) = \sum_{j=1}^\infty j c_j x^{j-1}$ . Suppose that the set is not linearly independent. Choose not-all-zero constants $\{\alpha_i\}_{i=1}^m$ and $\{\beta_{i1}, ..., \beta_{id}\}_{i=1}^m$ such that 
\begin{equation*}
    \vx \mapsto \sum_{i=1}^m \left( \alpha_i \sigma(\vw_i^\TT \vx) + \sum_{t=1}^d \beta_{it} \sigma'(\vw_i^\TT \vx)x_t \right) 
\end{equation*} 
is a zero map on $\sR^d$, where $x_t$ denotes the $t$-th component of input. For $k, j, i \in [d]$, define the sets 
\begin{align*}
    A_{k,j} &:= \{\vx \in \sR^d | \left< \vx, \vw_k \pm \vw_j \right> = 0\} \\ 
    B_i     &:= \{\vx \in \sR^d | \left<\vx, \vw_i \right> = 0\}. 
\end{align*} 
Clearly, each $A_{k,j}$ is the union of two linear subspaces of dimension $(d-1)$, while each $B_i$ is a possibly empty affine subspace of dimension $(d-1)$. Thus, 
\begin{equation*}
    E := \left( \cup_{1\leqslant k,j \leqslant m} A_{k,j} \right) \cup \left( \cup_{i=1}^d B_i \right)
\end{equation*} 
has $\fL^{d}$ Lebesgue measure zero. Let $\ve \in \sR^d \backslash E$. Denote $p_i := \left< \vw_i, \ve \right>$ for each $i \in [m]$. Since $p_i \neq p_j$ and $p_i + p_j \neq 0$ whenever $i \neq j$, we can, without loss of generality, assume that $|p_1| > |p_2| > ... > |p_m|>0$. For any sufficiently small $\vep$ and any $i, t$ we have 
\begin{align*}
    \sigma(\vw_i^\TT (\vep \ve)) &= \sum_{j=0}^\infty (c_j p_i^j) \vep^j, \\ 
    \sigma'(\vw_i^\TT (\vep\ve))(\vep\ve)_t &= e_t \sum_{j=1}^\infty (j c_j p_i^{j-1}) \vep^j.  
\end{align*}
Thus, for sufficiently small $\vep$, 
\begin{equation}\label{linear independence of activations eq1}
\begin{aligned}
    \sum_{i=1}^m \left( \alpha_i \sigma(\vw_i^\TT(\vep\ve)) + \sum_{t=1}^d \beta_{it} \sigma'(\vw_i^\TT (\vep\ve))(\vep\ve)_t\right) 
    &= \left(\sum_{i=1}^m \alpha_i \right) c_0 + \sum_{j=1}^\infty c_j \sum_{i=1}^m \left( \alpha_i + \frac{1}{p_i}\sum_{t=1}^d j\beta_{it} e_t \right) p_i^j \vep^j \\
    &= 0.
\end{aligned}
\end{equation}
We have $c_j \sum_{i=1}^m \left( \alpha_i + \frac{1}{p_i}\sum_{t=1}^d j\beta_{it} e_t \right) p_i^j = 0$ for all $j \in \sN$. In particular, for any $j \geq 2$, since $n_j \geq 1$ and $c_{n_j} \neq 0$, we have $\sum_{i=1}^m \left( \alpha_i + \frac{1}{p_i} \sum_{t=1}^d n_j\beta_{it} e_t \right) p_i^{n_j} = 0$, which yields 
\begin{equation*}
    \alpha_1 + \frac{1}{p_1}\sum_{t=1}^d n_j\beta_{1t} e_t = - \sum_{i=2}^m \left( \alpha_i + \frac{1}{p_i}\sum_{t=1}^d n_j\beta_{it} e_t \right) \frac{p_i^{n_j}}{p_1^{n_j}}. 
\end{equation*}
If $m = 1$, by taking limits $j\to \infty$, we have $\alpha_1 = \sum_{t=1}^d \beta_{1t}e_t = 0$.

Otherwise, since $|p_1| > |p_i|$ for any $2 \leqslant i \leqslant m$, it follows that, by taking limits $j\to \infty$, 
\begin{equation*}
     \lim_{j\rightarrow \infty}\left(\alpha_1 + \frac{1}{p_1}\sum_{t=1}^d n_j \beta_{1t} e_t \right) = \lim_{j\rightarrow \infty}- \sum_{i=2}^m \left( \alpha_i + \frac{1}{p_i}\sum_{t=1}^d n_j\beta_{it} e_t \right) \frac{p_i^{n_j}}{p_1^{n_j}} = 0. 
\end{equation*}
Thus, we also have $\alpha_1 = \sum_{t=1}^d \beta_{1t}e_t = 0$. For $m > 2$, we may rewrite Eq. \eqref{linear independence of activations eq1} as 
\begin{equation*}
    \alpha_2 +\frac{1}{p_2} \sum_{t=1}^d n_j \beta_{2t} e_t = - \sum_{i=3}^m \left( \alpha_i + \frac{1}{p_i}\sum_{t=1}^d n_j \beta_{it} e_t \right) \frac{p_i^{n_j}}{p_2^{n_j}}
\end{equation*} 
for each $j \geq 2$, and take limits as we do above to deduce that $\alpha_2 + \frac{1}{p_2}\sum_{t=1}^d n_j \beta_{2t} e_t = 0$. By repeating this procedure for  at most $m$ times, we conclude that $\alpha_i + \frac{1}{p_i}\sum_{t=1}^d n_j \beta_{it} e_t = 0$ for all $i \in [m]$. Then, $\alpha_i = \sum_{t=1}^d \beta_{it}e_t = 0$ for any $i\in [m]$. For each $i$, $\sum_{t=1}^d \beta_{it}e_t'$ is a linear function of $\ve'$ on the open set $\sR^d\backslash E$ which vanishes on a neighborhood of $\ve$, we must have $\alpha_i = \beta_{it} = 0$ for any $i \in [m], t \in [d]$. Therefore, $\{\sigma(\vw_i^\TT \vx), \sigma'(\vw_i^\TT \vx)x_1, ..., \sigma'(\vw_i^\TT \vx)x_d\}_{i=1}^m$ must be a linearly independent set. 
\end{proof}

\begin{cor}[model rank estimate for two-layer NNs]
Let $\sigma = \tanh$. Given $d \in \sN$, weights $\vw_1, ..., \vw_m \in \sR^d$, $a_1, ..., a_m \in \sR$, we have 
\begin{equation*}
    \rdim(\rspan \{\sigma(\vw_i^{\TT} \vx), a_i \sigma'(\vw_i^{\TT} \vx)x_1, ..., a_i \sigma'(\vw_i^{\TT} \vx)x_d \}_{i=1}^m)=m_{\vw} + m_{a} d,
\end{equation*} 
where $m_{\vw}=\frac{1}{2}|\{\vw_i,-\vw_i|\vw_i\neq \vzero, i\in[m]\}|$ indicating the number of independent neurons, $m_{a}=\frac{1}{2}|\{\vw_i,-\vw_i|\vw_i\neq \vzero, a_i\neq 0, i\in[m]\}| + |\{\vw_i|\vw_i= \vzero, a_i\neq 0, i\in[m]\}|$ indicating the number of independent effective neurons. Here, $|\cdot|$ is the cardinality  of a set, i.e., number of different elements in a set. 
\label{cor:dimension}
\end{cor}
\begin{proof}
Note that $\sigma = \tanh$ is analytic and $\sigma^{(2n+1)}(0) \neq 0$ for all $n$. Because $\tanh$ is an odd function, we have $\tanh(x) = -\tanh(-x)$ and $\tanh(0) = 0$. Therefore, given $\vw_i, \vw_j \neq \vzero$ with $\vw_i = \pm \vw_j$, $\rspan \{\sigma(\vw_i^\TT x)\} = \rspan\{\sigma(\vw_j^\TT x)\}$ and $\rspan \{\sigma'(\vw_i^\TT x)x_1, ..., \sigma'(\vw_i^\TT x)x_d\} = \rspan \{\sigma'(\vw_j^\TT x)x_1, ..., \sigma'(\vw_j^\TT x)x_d\}$. Since there are $m_{\vw}$ different non-zero weights, by Proposition~\ref{appendix prop: linear independence of activations} we have 
\begin{equation*}
    \dim \left( \rspan \{\sigma(\vw_i^\TT x)\}_{i=1}^m \right)= m_{\vw}. 
\end{equation*}
Furthermore, note that 
\begin{equation*}
    \rspan \{a_i \sigma'(\vw_i^\TT x)x_1, ..., a_i \sigma'(\vw_i^\TT x)x_d \}_{i=1}^m = \rspan \{\sigma'(\vw_i^\TT x)x_1, ..., \sigma'(\vw_i^\TT x)x_d: a_i \neq 0, i \in [m]\}. 
\end{equation*}
Thus, by Proposition~\ref{appendix prop: linear independence of activations}, 
\begin{align*}
    &\,\,\,\,\,\,\,\dim \left( \rspan \{\sigma'(\vw_i^\TT x)x_1, ..., \sigma'(\vw_i^\TT x)x_d: \vw_i \neq \vzero, a_i \neq 0, i \in [m]\} \right) \\ 
    &= \frac{1}{2}|\{\vw_i,-\vw_i|\vw_i\neq \vzero, a_i\neq 0, i\in[m]\}| \cdot d. 
\end{align*}
Now suppose that $\vw_j = \vzero \in \sR^d$ for some $j \in [m]$. Since $\sigma'(\vw^\TT x) = \sigma'(0) \neq 0$ for all $x \in \sR^d$, 
\begin{equation*}
    \rspan \{\sigma'(\vw_j^\TT x)x_1, ..., \sigma'(\vw_j^\TT x)x_d\} = \rspan \{x_1, ..., x_d\} 
\end{equation*} 
which consists only of linear functions. By the nonlinearity of $\tanh$, we conclude that 
\begin{equation*}
    \dim \left( \rspan \{\sigma'(\vw_i^\TT x)x_1, ..., \sigma'(\vw_i^\TT x)x_d\} \right) = m_a d 
\end{equation*} 
and thus $\dim \left( \rspan \{\sigma(\vw_i^\TT x), \sigma'(\vw_i^\TT x)x_1, ..., \sigma'(\vw_i^\TT x)x_d \} \right) = m_{\vw} + m_a d$ as desired. 
\end{proof}

The above corollary directly gives rise to the following result.
\begin{cor}
\label{cor:app_DNN}
Let $\sigma = \tanh$. Given distinct weights $\vw_1, ..., \vw_m \in \sR^d\backslash\{\boldsymbol{0}\}$ satisfying $\vw_k\neq\pm \vw_j$ for $k\neq j$ , and $a_1, ..., a_m \in \sR\backslash\{0\}$, we have 
\begin{equation*}
    \rdim(\rspan \{\sigma(\vw_i^{\TT} \vx), a_i \sigma'(\vw_i^{\TT} \vx)x_1, ..., a_i \sigma'(\vw_i^{\TT} \vx)x_d \}_{i=1}^m)=m(d+1).
\end{equation*}
\end{cor}

Next we consider the estimate of the model rank for convolutional neural networks (CNNs) which are widely used in practice. Here we consider the case where the input has two-dimensional indices, which is the most general case for the image input. The following two propositions can be directly generalized to the model rank estimate of CNNs with an input of one index dimension in the main text.

\begin{prop}[model rank estimate for CNNs (with weight sharing)]
    Given $m \in \sN$, $d \in \sN$ and $s \in [d]$. For any $l \in [m]$, let $\mK_l$ be a $(s \times s)$ matrix. 
    Consider CNNs with stride = 1. For a tanh-CNN $f_{\vtheta}$ with weight sharing, 
    \begin{equation*}
        f_{\vtheta} (\mI) = \sum_{l=1}^m \sum_{i,j=1}^{d+1-s} a_{lij} \tanh \left(\sum_{\alpha, \beta} I_{i+s-\alpha, j+s-\beta} K_{l;\alpha, \beta}\right), \mI \in \sR^{d\times d}, 
    \end{equation*}
    for $\mK_l\neq \vzero, 1\leq l\leq m$, its model rank at $\vtheta = (a_{lij}, \mK_{l})_{l, i, j}$ is $m_a s^2 + m_K(d+1-s)^2$, where $m_{K}=\frac{1}{2}|\{\mK_l,-\mK_l|l\in[m]\}|$ indicating the number of independent kernels, $m_{a}=\frac{1}{2}\sum_{\mK\in \fK}\operatorname{dim}(\rspan\{a_{l,:,:}\}_{l\in h(\mK)})$ indicating the number of independent effective neurons. Here $\fK = \{\mK_l, -\mK_l|l\in [m]\}$, $h$ is a function over $\fK$ s.t. for each $\mK\in \fK, h(\mK) = \{l|l\in [m], \mK_l = \pm\mK\}$. $|\cdot|$ is the cardinality  of a set, i.e., number of different elements in a set, and $a_{l,:,:}$ denotes the $(d+1-s) \times (d+1-s)$ matrix whose entries are $a_{lij}$'s.
    \label{prop:app_CNN}
\end{prop}
\begin{remark}
    In presence of zero-kernels at a parameter point $\vtheta$, i.e., $\mK_l=\mzero$ for certain $l$'s, the model rank is obviously no less than that at a parameter point $\vtheta'$ obtained by replacing $a_{lij}$ by $0$ for these $l$'s and all $i,j$'s in $\vtheta$.  Because we always have $f(\cdot;\vtheta)=f(\cdot;\vtheta')$ and the model rank at $\vtheta'$ is always the same as that in a narrower NN with all zero-kernels removed, establishing the rank estimate for parameter points with nonzero-kernels is sufficient for the rank estimate  over the mode function space.
\end{remark}

\begin{proof}
    We first consider the case in which $\mK_l \pm \mK_{l'} \neq \mzero$ for any distinct $l,l' \in [m]$. Let $\sigma = \tanh$. In this case the model rank is the dimension of the following function space (with respect to variable $\mI\in\sR^{d\times d}$) 
\begin{align*}
    &\,\,\,\,\,\,\,\rspan \left\{\frac{\partial f_{\vtheta}}{\partial a_{lij}} , \frac{\partial f_{\vtheta}}{\partial K_{l; \alpha, \beta}} \right\} \\ 
    &= \rspan \left\{ \sigma \left(\sum_{\alpha', \beta'} I_{i+s-\alpha', j+s-\beta'}K_{l;\alpha', \beta'} \right), \right. \\ 
    &\,\,\,\,\,\,\,\left.\sum_{i',j'=1}^{d+1-s} a_{li'j'} \sigma'\left(\sum_{\alpha', \beta'} I_{i'+s-\alpha', j'+s-\beta'}K_{l;\alpha', \beta'}\right) I_{i'+s-\alpha, j'+s-\beta} \right\}_{l, i, j,\alpha, \beta}, 
\end{align*}
where $l \in [m]$ and $\alpha, \beta \in [s]$. Next, we prove by contradiction that the set of functions 
\begin{equation*}
    \left\{ \sum_{i,j=1}^{d+1-s} a_{lij} \sigma' \left(\sum_{\alpha', \beta'} I_{i+s-\alpha', j+s-\beta'}K_{l;\alpha', \beta'} \right) I_{i+s-\alpha, j+s-\beta} \right\}_{l \in [m], \alpha,\beta \in [s]}
\end{equation*} 
are linearly independent.  If they are not linearly independent, there exist not all zero constants $\zeta_{l11}, ..., \zeta_{lss}$ for $l\in[m]$, such that 
\begin{equation*}
    \sum_{l=1}^m \sum_{\alpha, \beta=1}^s \zeta_{l\alpha\beta} \sum_{i,j=1}^{d+1-s} a_{lij} \sigma' \left(\sum_{\alpha', \beta'} I_{i+s-\alpha', j+s-\beta'}K_{l;\alpha', \beta'} \right) I_{i+s-\alpha, j+s-\beta}= 0, 
\end{equation*} 
which implies that the set of functions 
\begin{equation*}
    \left\{ a_{lij} \sigma'\left( \sum_{\alpha', \beta'} I_{i+s-\alpha', j+s-\beta'}K_{l;\alpha', \beta'}\right) I_{i+s-\alpha, j+s-\beta}\right\}_{l,i,j,\alpha,\beta}
\end{equation*} 
are linearly dependent, contradicting Proposition \ref{appendix prop: linear independence of activations}. Moroever, if $a_{lij} = 0$ for any $l \in [m]$ and all $i,j \in \{1, ..., d+1-s\}$, 
\begin{equation*}
    \sum_{i,j=1}^{d+1-s} a_{lij} \sigma' \left(\sum_{\alpha', \beta'} I_{i+s-\alpha', j+s-\beta'}K_{l;\alpha', \beta'} \right) I_{i+s-\alpha, j+s-\beta}= 0
\end{equation*} 
for all $\alpha, \beta \in [s]$. Notice that two kernels with $\mK_l = \pm\mK_{l'}$ can be reduced to one while maintaining model rank if and only if the corresponding output weights $a_{l,:,:}$ and $a_{l',:,:}$ are linearly dependent. Then, similar to Corollary~\ref{appendix prop: linear independence of activations}, we conclude that the model rank is $m_a s^2 + m_K(d+1-s)^2$. 
\end{proof}

\begin{prop}[model rank estimate for CNNs without weight sharing]
Given $m \in \sN$, $d \in \sN$ and $s \in [d]$. For any $l \in [m]$ and $i,j \in [d+1-s]$, let $\mK_{lij}$ be a $(s \times s)$ matrices. 
Consider CNNs with stride = 1. 
For a $\tanh$ CNN $f_{\vtheta}$ without weight sharing, 
\begin{equation*}
    f_{\vtheta} (\mI) = \sum_{l=1}^m \sum_{i,j=1}^{d+1-s} a_{lij} \tanh\left(\sum_{\alpha, \beta} I_{i+s-\alpha, j+s-\beta}K_{lij; \alpha, \beta}\right), \mI \in \sR^{d\times d}, 
\end{equation*}
for $\mK_{lij}\neq \vzero, l\in[m], i,j \in [d+1-s]$, its model rank at $\vtheta = (a_{lij}, \mK_{lij})_{l, i, j}$ is $m_a s^2 + m_K$, where $m_{K}=\frac{1}{2}|\{p(\mK_{lij}),-p(\mK_{lij})|l\in[m], i,j\in [d+1-s]\}|$ indicating the number of independent kernels, $m_{a}=\frac{1}{2}|\{p(\mK_{lij}),-p(\mK_{lij})|l\in[m], i,j\in [d+1-s], a_{lij}\neq 0\}|$ indicating the number of independent effective neurons. Here $p$ is the padding function over kernels, i.e., for each ($s\times s$) kernel $\mK_{lij}$, $p(\mK_{lij}) \in \sR^{d\times d}$ s.t. $p(\mK_{lij})[i:i+s-1, j:j+s-1] = \mK_{lij}$ and the other elements of $p(\mK_{lij})$ are zero. $|\cdot|$ is the cardinality  of a set, i.e., number of different elements in a set. 
\label{prop:app_CNN_without_weight_sharing}
\end{prop}

\begin{remark}
    Similar to CNNs (with weight sharing), for CNNs without weight sharing, establishing the rank estimate for parameter points with nonzero-kernels is sufficient for the rank estimate over the mode function space.
    
\end{remark}

\begin{proof}
Let $\sigma=\tanh$. The model rank is the dimension of the following function space 
\begin{align*}
    &\,\,\,\,\,\,\,\rspan \left\{\frac{\partial f_{\vtheta}}{\partial a_{lij}} , \frac{\partial f_{\vtheta}}{\partial K_{lij; \alpha, \beta}} \right\}_{l,i,j,\alpha,\beta} \\ 
    &= \rspan \left\{ \sigma \left(\sum_{\alpha', \beta'} I_{i+s-\alpha', j+s-\beta'} K_{lij;\alpha', \beta'} \right), \right.\\
    &\quad \left.a_{lij} \sigma' \left(\sum_{\alpha', \beta'} I_{i+s-\alpha', j+s-\beta'} K_{lij;\alpha', \beta'} \right) I_{i+s-\alpha, j+s-\beta}\right\}_{l,i,j,\alpha, \beta}, 
\end{align*}
where $l \in [m]$, $1 \leqslant i, j \leqslant d+1-s$, and $\alpha, \beta \in [s]$. Also note that if $a_{lij} = 0$ for some $l \in [m]$ and $i,j \in \{1, ..., d+1-s\}$, then 
\begin{equation*}
    a_{lij} \sigma' \left(\sum_{\alpha', \beta'} I_{i+s-\alpha', j+s-\beta'} K_{lij;\alpha', \beta'} \right) I_{i+s-\alpha, j+s-\beta}= 0
\end{equation*}
for all $\alpha, \beta \in [s]$. It follows from Proposition \ref{appendix prop: linear independence of activations} that this space has dimension $m_a s^2 + m_K$. 
\end{proof}

\begin{remark}
    In particular, for a target function $f^*\in \mathcal{F}_m^{\mathrm{CNN}}\backslash\mathcal{F}_{m-1}^{\mathrm{CNN}}$, the model rank of a equivalent CNNs model without weight sharing is $m_a s^2+ m_K=m(s^2+1)(d+1-s)^2 - s^2m_{\mathrm{null}}$, where $m_{\mathrm{null}}=|\{a_{lij}|a_{lij}=0\}|$ is a variable counting the number of ineffective neurons in the target function.    
\end{remark}

\subsubsection{Two-layer fully-connected neural networks\label{appsec:FCNN_estimate}}
\textbf{Two-layer tanh-NN:} $f_{\boldsymbol{\theta}}(\boldsymbol{x}) = \sum_{i=1}^{m}a_i\sigma(\boldsymbol{w}_i^\TT \boldsymbol{x}), \boldsymbol{x}\in \mathbb{R}^d, \boldsymbol{\theta} = (a_i, \boldsymbol{w}_i)_{i=1}^m, \sigma = \tanh.$

\textbf{Step 1:} Stratify the parameter space into different model rank levels to obtain the rank hierarchy over the parameter space.

Given any parameter point $\vtheta = (a_i, \vw_i)_{i=1}^{m}$. Consider the tangent space 
$$\operatorname{span}\left\{\sigma(\boldsymbol{w}_i^\TT\boldsymbol{x} ), a_i\sigma'(\boldsymbol{w}_i^\TT \boldsymbol{x})x_1, \cdots, a_i\sigma'(\boldsymbol{w}_i^\TT \boldsymbol{x})x_d\right\}_{i=1}^{m}.$$
By Corollary~\ref{cor:dimension}, the dimension of tangent space is $m_{a} d+m_{\vw}$, where $m_{\vw}=\frac{1}{2}|\{\vw_j,-\vw_j|j\in[m],\vw_j\neq \vzero\}|$ indicating the number of independent neurons, $m_{a}=\frac{1}{2}|\{\vw_j,-\vw_j|j\in[m], \vw_j\neq \vzero, a_j\neq 0\}| + |\{\vw_j|j\in[m], \vw_j= \vzero, a_j\neq 0\}|$ indicating the number of independent effective neurons. Here, $|\cdot|$ is the cardinality  of a set, i.e., number of different elements in a set.  

\textbf{Step 2:} Stratify the model function space into different model rank levels to obtain the rank hierarchy over the model function space.

Given any target function $f^*$ that can be recovered by a two-layer NN with width $m$. Without loss of generality, let $f^* = f_{\vtheta^*} := \sum_{i=1}^k a^*_i\sigma(\boldsymbol{w}_i^{*\TT}\boldsymbol{x}), a_i^*\neq0, \vw^*_i\neq\vzero, \vw^*_i\neq\pm\vw^*_j, 1\leq k \leq m, \vtheta^* = (a_i^*, \vw_i^*)_{i=1}^{k}$. By Proposition~\ref{appendix prop: linear independence of activations}, the set $\{\sigma(\boldsymbol{w}_i^\TT\boldsymbol{x} ), a_i\sigma'(\boldsymbol{w}_i^\TT \boldsymbol{x})x_1, \cdots, a_i\sigma'(\boldsymbol{w}_i^\TT \boldsymbol{x})x_d\}_{i=1}^{k}$ is linearly independent and $\rrank_{f_{\vtheta}}(\vtheta^*) = k(d+1)$. By definition, the model rank of $f^*$ is the minimal model rank among all parameters recovering $f^*$. Suppose there exists a NN $f_{\vtheta}= \sum_{i=1}^q a_i\sigma(\boldsymbol{w}_i^{\TT}\boldsymbol{x}), a_i\neq0, \vw_i\neq\vzero, \vw_i\neq\pm\vw_j$ for $i\neq j$ that can recover $f^*$, and $\{\sigma(\boldsymbol{w}_i^\TT\boldsymbol{x} ), a_i\sigma'(\boldsymbol{w}_i^\TT \boldsymbol{x})x_1, \cdots, a_i\sigma'(\boldsymbol{w}_i^\TT \boldsymbol{x})x_d\}_{i=1}^{q}=q(d+1)<k(d+1)$, then we have $q<k$. Since $\{\sigma(\boldsymbol{w}_i^\TT\boldsymbol{x})\}_{i=1}^k$ is linearly independent and $\rdim(\rspan\{\sigma(\boldsymbol{w}_i^\TT\boldsymbol{x} )\}_{i=1}^{q})\leq q<k$, this contradicts $f_{\vtheta}$ being able to recover $f^*$. Therefore, $\rrank_{f_{\vtheta}}(\vtheta)\geq k(d+1)$ and $\rrank_{f_{\vtheta}}(\vtheta)$ attains its lower bound $k(d+1)$ at $\vtheta^*$. Thus $\rrank_{f_{\vtheta}}(f^*)=k(d+1)$. Then, the two-layer NN model possesses the rank levels $\{k(d+1)|k\in[m]\}$ over its function space, each of which is occupied by $\{f: \sR^{d}\rightarrow \sR|f = f_{\vtheta} := \sum_{i=1}^k a_i\sigma(\boldsymbol{w}_i^{\TT}\boldsymbol{x}), a_i\neq0, \vw_i\neq\vzero, \vw_i\neq\pm\vw_j\}$ as illustrated in Table \ref{table:FFN} in the main text.

Remark that the above analysis serves as a proof of the following proposition.
\begin{prop}[rank hierarchy of two-layer tanh-NN] 
Given a two-layer NN $$f_{\boldsymbol{\theta}}(\boldsymbol{x}) = \sum_{i=1}^{m}a_i\tanh(\boldsymbol{w}_i^\TT \boldsymbol{x}), \boldsymbol{x}\in \mathbb{R}^d, \boldsymbol{\theta} = (a_i, \boldsymbol{w}_i)_{i=1}^m,$$
for any target function $f^*$ that can be recovered by a two-layer NN with width $m$, say $f^* = f_{\vtheta^*} := \sum_{i=1}^k a^*_i\sigma(\boldsymbol{w}_i^{*\TT}\boldsymbol{x}), a_i^*\neq0, \vw^*_i\neq\vzero, \vw^*_i\neq\pm\vw^*_j, 1\leq k \leq m, \vtheta^* = (a_i^*, \vw_i^*)_{i=1}^{k}$ , we have model rank
$$\rrank_{f_{\vtheta}}(f^*)=k(d+1).$$
\end{prop}

\subsubsection{Two-layer convolutional neural networks \label{appsec:CNN_estimate}}
\textbf{Two-layer tanh-CNN with weight sharing.} Consider the $2$-layer width-$m$ tanh convolution neural networks without weight sharing. Given $m \in \sN$, $d \in \sN$ and $s \in [d]$. For any $l \in [m]$ and $i,j \in \{1, ..., d+1-s\}$, let $\mK_{l}$ be a $(s \times s)$ convolutional kernel. 
For a 2-d input $I$, consider tanh-CNNs with stride = 1:
\begin{align*}
    f_{\vtheta} (I) = \sum_{l=1}^m \sum_{i,j=1}^{d+1-s} a_{lij} \sigma\left(\sum_{\alpha, \beta} I_{i+s-\alpha, j+s-\beta}K_{l; \alpha, \beta}\right), \quad \mI \in \sR^{d\times d}, \sigma = \tanh
\end{align*}

\textbf{Step 1:} Stratify the parameter space into different model rank levels to obtain the rank hierarchy over the parameter space.

Given any parameter point $\vtheta = (a_{lij}, \mK_{l})_{l, i, j} (\mK_l\neq \vzero)$. Consider the tangent space 
\begin{align*}
    &\rspan \left\{ \sigma \left(\sum_{\alpha', \beta'} I_{i+s-\alpha', j+s-\beta'}K_{l;\alpha', \beta'} \right), \right. \\ 
    &\left.\sum_{i',j'=s+1}^{d+1} a_{i'j'l} \sigma'\left(\sum_{\alpha', \beta'} I_{i'+s-\alpha', j'+s-\beta'}K_{l;\alpha', \beta'}\right) I_{i'+s-\alpha, j'+s-\beta} \right\}_{l,i, j,\alpha, \beta}. 
\end{align*}
By Proposition~\ref{prop:app_CNN}, the dimension of tangent space is $m_a s^2 + m_K(d+1-s)^2$, where $m_{K}=\frac{1}{2}|\{\mK_l,-\mK_l|l\in[m]\}|$ indicating the number of independent kernels, $m_{a}=\frac{1}{2}\sum_{\mK\in \fK}\operatorname{dim}(\rspan\{a_{l,:,:}\}_{l\in h(\mK)})$ indicating the number of independent effective neurons. Here $\fK = \{\mK_l, -\mK_l|l\in [m]\}$ and $h$ is a function over $\fK$ s.t. for each $\mK\in \fK, h(\mK) = \{l|l\in [m], \mK_l = \pm\mK\}$. $|\cdot|$ is the cardinality  of a set, i.e., number of different elements in a set.

\textbf{Step 2:} Stratify the model function space into different model rank levels to obtain the rank hierarchy over the model function space.

Given any target function $f^*$ that can be recovered by a two-layer NN with width $m$. Without loss of generality, let $f^* = f_{\vtheta^*} := \sum_{l=1}^{k}\sum_{i,j=1}^{d+1-s} a^*_{lij} \sigma\left(\sum_{\alpha, \beta} I_{i+s-\alpha, j+s-\beta}K^*_{l; \alpha, \beta}\right)$, $\mK^*_{l}\neq \boldsymbol{0}, \mK^*_{l}\neq \pm \mK^*_{l'}$ for any $l\neq l', \forall l,  \exists a^*_{lij}\neq 0, 1\leq k \leq m, \vtheta^* = (a^*_{lij}, \mK^*_{l})_{l, i, j}$. By Proposition~\ref{appendix prop: linear independence of activations}, $\rrank_{f_{\vtheta}}(\vtheta^*) = k(s^2 + (d+1-s)^2)$. By definition, the model rank of $f^*$ is the minimal model rank among all parameters recovering $f^*$. Suppose there exists a NN $f_{\vtheta}= \sum_{l=1}^{q}\sum_{i,j=1}^{d+1-s} a^*_{lij} \sigma\left(\sum_{\alpha, \beta} I_{i+s-\alpha, j+s-\beta}K^*_{l; \alpha, \beta}\right)$, $\mK^*_{l}\neq \boldsymbol{0}, \mK^*_{l}\neq \pm \mK^*_{l'}$ for any $l\neq l', \forall l,  \exists a^*_{lij}\neq 0$ that can recover $f^*$ and the dimension of tangent space $q(s^2+(d+1-s)^2)<k(s^2+(d+1-s)^2)$, then we have $q<k$. Since $\{\sigma\left(\sum_{\alpha, \beta} I_{i+s-\alpha, j+s-\beta}K^*_{l; \alpha, \beta}\right)\}_{l=1}^k$ is linearly independent and $\rdim(\rspan\{\sigma\left(\sum_{\alpha, \beta} I_{i+s-\alpha, j+s-\beta}K^*_{l; \alpha, \beta}\right)\}_{l=1}^{q})\leq q<k$, this contradicts $f_{\vtheta}$ being able to recover $f^*$. Therefore, $\rrank_{f_{\vtheta}}(\vtheta)\geq k(s^2+(d+1-s)^2)$ and $\rrank_{f_{\vtheta}}(\vtheta)$ attains its lower bound $k(s^2+(d+1-s)^2)$ at $\vtheta^*$. Thus $\rrank_{f_{\vtheta}}(f^*)=k(s^2+(d+1-s)^2)$. Then, the two-layer CNN model possesses the rank levels $\{k(s^2+(d+1-s)^2)|k\in[m]\}$ over its function space, each of which is occupied by $\{f: \sR^{d}\rightarrow \sR|f = f_{\vtheta} := \sum_{l=1}^{k}\sum_{i,j=1}^{d+1-s} a_{lij} \sigma\left(\sum_{\alpha, \beta} I_{i+s-\alpha, j+s-\beta}K_{l; \alpha, \beta}\right)\}$ as illustrated in Table \ref{table:CNN-with-weight sharing}.

\begin{table}[htb]
\caption{The rank hierarchy of two-layer width-$m$ tanh-CNN (with weight sharing).}
\centering
\renewcommand{\arraystretch}{1.8} 
\begin{tabular}{|c|c|}
\hline
\multicolumn{2}{|c|}{model: $f_{\vtheta} (I) = \sum_{l=1}^m \sum_{i,j=1}^{d+1-s} a_{lij} \sigma\left(\sum_{\alpha, \beta} I_{i+s-\alpha, j+s-\beta}K_{l; \alpha, \beta}\right), \mI \in \sR^{d\times d}, \vtheta = (a_{lij}, \mK_{l})_{l, i, j}$}                                         \\ \hline
$\rrank_{f_{\vtheta}}(f^*)$                   & \multicolumn{1}{c|}{$f^*$}                                                               \\ \hline
$0$                                           & \multicolumn{1}{c|}{$0$}                         \\ \hline
$s^2 +(d+1-s)^2$                                        & \multicolumn{1}{c|}{\makecell[c]{$\mathcal{F}^{\mathrm{CNN}}_1\backslash \{0\}: \{\sum_{i,j=1}^{d+1-s} a^*_{ij} \sigma\left(\sum_{\alpha, \beta} I_{i+s-\alpha, j+s-\beta}K^*_{1; \alpha, \beta}\right)|$\\ $\mK_1^*\neq \boldsymbol{0}, \exists i,j, a^*_{ij}\neq 0\}$}} \\ \hline
$\vdots$                                      & \multicolumn{1}{c|}{$\vdots$}                                                          \\ \hline
$k(s^2 +(d+1-s)^2)$                                    & \multicolumn{1}{c|}{\makecell[c]{$\mathcal{F}^{\mathrm{CNN}}_k\backslash \mathcal{F}^{CNN}_{k-1}: \{\sum_{l=1}^{k}\sum_{i,j=1}^{d+1-s} a^*_{lij} \sigma\left(\sum_{\alpha, \beta} I_{i+s-\alpha, j+s-\beta}K^*_{l; \alpha, \beta}\right)|$\\ $\mK^*_{l}\neq \boldsymbol{0}, \mK^*_{l}\neq \pm \mK^*_{l'}$ for any $l\neq l', \forall l,  \exists a^*_{lij}\neq 0\}$}} \\ \hline
$\vdots$                                      & \multicolumn{1}{c|}{$\vdots$}                                                         \\ \hline
$m(s^2 +(d+1-s)^2)$                                        & \multicolumn{1}{c|}{\makecell[c]{$\mathcal{F}^{\mathrm{CNN}}_m\backslash \mathcal{F}^{CNN}_{m-1}: \{\sum_{l=1}^{m}\sum_{i,j=1}^{d+1-s} a^*_{lij} \sigma\left(\sum_{\alpha, \beta} I_{i+s-\alpha, j+s-\beta}K^*_{l; \alpha, \beta}\right)|$\\ $\mK^*_{l}\neq \boldsymbol{0}, \mK^*_{l}\neq \pm \mK^*_{l'}$ for any $l\neq l', \forall l,  \exists a^*_{lij}\neq 0\}$}} \\ \hline
\end{tabular}
\label{table:CNN-with-weight sharing}
\end{table}

\subsubsection{Details of architecture comparison\label{appsec:arc_comp}}
\begin{figure}[htbp]
    \centering
    \includegraphics[width=0.9\textwidth]{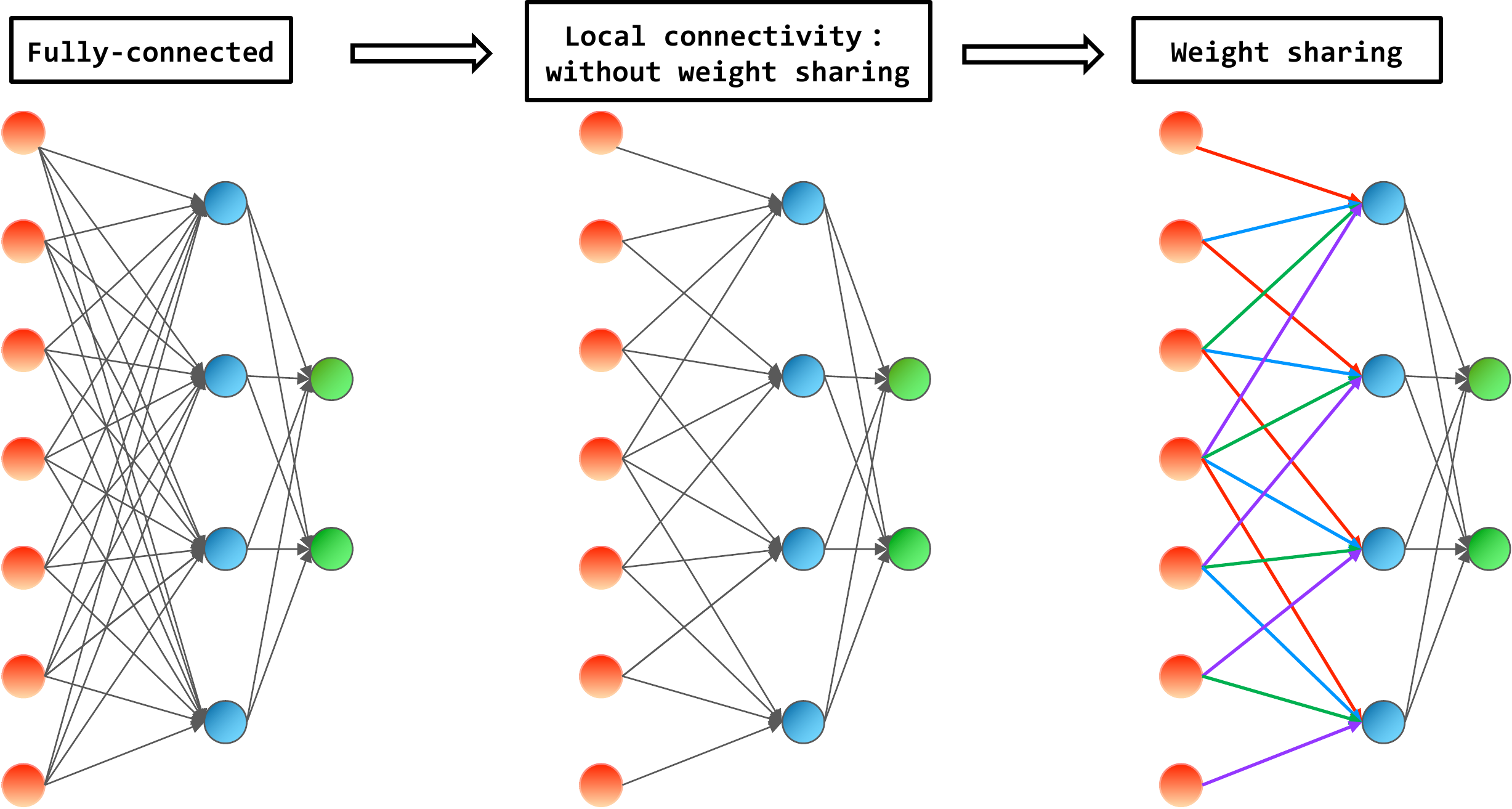}
    \caption{Illustration of architectures from fully-connected NN to CNN for comparison. \label{fig:comp}}
\end{figure}

On the basis of the above rank hierarchy for two-layer CNNs with weight sharing, we further exhibit in Table. \ref{table:model_comparing} the model rank in other architectures such as CNNs without weight sharing and fully-connected NNs illustrated in Fig. \ref{fig:comp}. Remark that the total size of hidden neurons $m(d+1-s)^2$ is fixed over different architectures. The model of CNN without weight sharing is introduced below.

\textbf{Two-layer tanh-CNN without weight sharing.} Consider the $2$-layer width-$m$ tanh convolution neural networks without weight sharing. Given $m \in \sN$, $d \in \sN$ and $s \in [d]$. For any $l \in [m]$ and $i,j \in \{1, ..., d+1-s\}$, let $\mK_{lij}$ be a $(s \times s)$ convolutional kernel. 
For a 2-d input $I$, consider tanh-CNNs with stride = 1:
\begin{align*}
    f_{\vtheta} (I) = \sum_{l=1}^m \sum_{i,j=1}^{d+1-s} a_{lij} \sigma\left(\sum_{\alpha, \beta} I_{i+s-\alpha, j+s-\beta}K_{lij; \alpha, \beta}\right), \quad \mI \in \sR^{d\times d}, \sigma = \tanh
\end{align*}

\begin{table}[htb]
\caption{The rank hierarchy for two-layer tanh-CNN with $m$-kernels of size $s\times s$ and stride $1$. The input $\vx\in\sR^{d\times d}$. For functions in each function set over the rank hierarchy, we also present their model rank in the corresponding CNN without weight sharing and the corresponding DNN. Here $m_{\mathrm{null}}=|\{a_{lij}|a_{lij}=0\}|$ counts the number of zero output weights in the target function. \label{table:model_comparing}}
\centering
\renewcommand{\arraystretch}{1.6} 
\begin{tabular}{|c|c|c|c|}
\hline
$f^*$                                                           & CNN & CNN without weight sharing                    & Fully-connected NN                            \\ \hline
$0$                                                             & $0$                     & $0$                                           & $0$                                           \\ \hline
$\mathcal{F}_1^{\mathrm{CNN}}\backslash\{0\}$                   & $s^2 + (d+1-s)^2$       & $(s^2+1)(d+1-s)^2-s^2m_{\mathrm{null}}$  & $(d^2+1)(d+1-s)^2-d^2m_{\mathrm{null}}$  \\ \hline
$\vdots$                                                        & $\vdots$                & $\vdots$                                      & $\vdots$                                      \\ \hline
$\mathcal{F}_k^{\mathrm{CNN}}\backslash\mathcal{F}_{k-1}^{\mathrm{CNN}}$ & $k(s^2 + (d+1-s)^2)$    & $k(s^2+1)(d+1-s)^2-s^2m_{\mathrm{null}}$ & $k(d^2+1)(d+1-s)^2-d^2m_{\mathrm{null}}$ \\ \hline
$\vdots$                                                        & $\vdots$                & $\vdots$                                      & $\vdots$                                      \\ \hline
$\mathcal{F}_m^{\mathrm{CNN}}\backslash\mathcal{F}_{m-1}^{\mathrm{CNN}}$ & $m(s^2 + (d+1-s)^2)$    & $m(s^2+1)(d+1-s)^2-s^2m_{\mathrm{null}}$ & $m(d^2+1)(d+1-s)^2-d^2m_{\mathrm{null}}$ \\ \hline
\end{tabular}
\end{table}
By Table. \ref{table:model_comparing}, for a common data of $d=28$, suppose the target function can be recovered by a CNN model with width $m$, then the model rank for different NN architectures varies a lot from $685m$ (CNN with weight sharing)  to $6760m$ (CNN without weight sharing) to $530660m$ (fully-connected NN), indicating a huge difference in their target recovery/generalization performance with limit training data.

\section{Details of linear stability theory\label{appsec:LST}}

\begin{definition}[linear stability for recovery]
Given any differentiable model $f_{\vtheta}$ with model function space $\fF_{f_{\vtheta}}$, loss function $\ell(\cdot,\cdot)$, and training data $S=\{(\vx_i,y_i)\}_{i=1}^n$, \\
(i) a parameter point $\vtheta^*\in\sR^M$ is linearly stable if $f(\cdot;\vtheta^*)$ is the unique solution to 
\begin{equation}
    \mathrm{min}_{f\in\widetilde{\fP}_{\vtheta^*}}\frac{1}{n}\sum_{i=1}^{n}\ell(f(\vx_i),y_i), \label{appeq:minF}
\end{equation}
where the tangent function hyperplane $\widetilde{\fP}_{\vtheta^*}:=f(\cdot;\vtheta^*)+\fP_{\vtheta^*}=\{f(\cdot;\vtheta^*)+\va^\TT\nabla_{\vtheta} f(\cdot;\vtheta^*)|\va\in\sR^M\}$;\\
(ii) a function $f^*\in\fF_{f_{\vtheta}}$ is linearly stable if there exists a linearly stable parameter point $\vtheta'$ such that $f(\cdot;\vtheta')=f^*$.
\end{definition}

\begin{definition}[empirical tangent matrix and empirical model rank]
Given any differentiable model $f_{\vtheta}$ and training data $S=\{(\vx_i,y_i)\}_{i=1}^n$, at any parameter point $\vtheta^*$,  $\nabla_{\vtheta}f(\mX;\vtheta^*)=[\nabla_{\vtheta}f(\vx_1;\vtheta^*),\cdots, \nabla_{\vtheta}f(\vx_n;\vtheta^*)]$ is referred to as the empirical tangent matrix. Then the empirical model rank is defined as follows
$$\rrank_{S}(\vtheta^*)=\rrank(\nabla_{\vtheta}f(\mX;\vtheta^*)).$$
\end{definition}

\begin{assumption}\label{ass:loss}
The loss function $\ell: \sR \times \sR \to [0,\infty)$ is a continuously differentiable function satisfying $\ell(x, y) = 0$ if and only if $x = y$.  
\end{assumption}

\begin{lemma}[linear stability condition for recovery]\label{thm:LSC_app}
Given any differentiable model $f_{\vtheta}$ and training data $S=\{(\vx_i,y_i)\}_{i=1}^n$,
a global minimizer $\vtheta^*\in\sR^M$ satisfying $f(\vx_i;\vtheta^*)=y_i$ for all $i\in[n]$ is linearly stable if and only if $\rrank_{S}(\vtheta^*)=\rrank_{f_{\vtheta}}(\vtheta^*)$.
\end{lemma}

\begin{proof}
Let
\begin{equation*}
    \Tilde{R}_S(\va) = \frac{1}{n}\sum_{i=1}^n \ell(f(\vx_i, \vtheta^*) + \va^\TT \nabla_{\vtheta}f(\vx_i;\vtheta^*), y_i). 
\end{equation*}
Because $\ell(f(\vx_i, \vtheta^*), y_i) = 0$ for all $i \in [n]$, it follows that $\va$ is a global minimum of $\Tilde{R}_S$ if and only if $\va \in \ker (\nabla_{\vtheta}f(\mX;\vtheta^*)^\TT)=\{\nabla_{\vtheta}f(\mX;\vtheta^*)^\TT \va=\vzero|\va\in\sR^M\}$. Now if $\vtheta^*$ is linearly stable, because $f(\cdot,\vtheta^*)$ is the unique solution, for any $\va \in \sR^M$ such that $\Tilde{R}_S(\va) = 0$, we must have $\va \in \ker (\nabla_{\vtheta}f(\cdot;\vtheta^*)^\TT)=\{\nabla_{\vtheta}f(\cdot;\vtheta^*)^\TT \va=0|\va\in\sR^M\}$, thus $\ker (\nabla_{\vtheta}f(\mX;\vtheta^*)^\TT)\subseteq\ker (\nabla_{\vtheta}f(\cdot;\vtheta^*)^\TT)$. But since $\ker (\nabla_{\vtheta}f(\cdot;\vtheta^*)^\TT) \subseteq \ker (\nabla_{\vtheta}f(\mX;\vtheta^*)^\TT)$ based on the fact that zero function attains $0$ at any data point, we have $\ker (\nabla_{\vtheta}f(\cdot;\vtheta^*)^\TT) = \ker (\nabla_{\vtheta}f(\mX;\vtheta^*)^\TT)$. Because $\rrank_{S}(\vtheta^*)+\rdim(\ker (\nabla_{\vtheta}f(\mX;\vtheta^*)^\TT))=\rrank_{f_{\vtheta}}(\vtheta^*)+\rdim(\ker (\nabla_{\vtheta}f(\cdot;\vtheta^*)^\TT))\equiv M$, we obtain $\rrank_{S}(\vtheta^*)=\rrank_{f_{\vtheta}}(\vtheta^*)$.

Conversely, if $\rrank_{S}(\vtheta^*)=\rrank_{f_{\vtheta}}(\vtheta^*)$, then $\ker (\nabla_{\vtheta}f(\cdot;\vtheta^*)^\TT) = \ker (\nabla_{\vtheta}f(\mX;\vtheta^*)^\TT)$. Thus, for any $\va \in \sR^M$ with $\Tilde{R}_S(\va) = 0$, we have $\vtheta \in \ker (\nabla_{\vtheta}f(\mX;\vtheta^*)^\TT) = \ker (\nabla_{\vtheta}f(\cdot;\vtheta^*)^\TT)$. Therefore, $f(\cdot, \vtheta^*)$ is the unique solution in its tangent function hyperplane, i.e., $\vtheta^*$ is linearly stable. 
\end{proof}

\begin{lemma}\label{lem:zero-measure}
Given $m$ linearly independent analytic functions $\phi_1(\vx),\cdots,\phi_m(\vx)$ with $\phi_i:\sR^d\to\sR$ for all $i\in[m]$, $\rrank(\mPhi(\mX))= m$ a.e. with respect to   $\fL^{d\times m}$  Lebesgue measure, where 
    \begin{equation*}
    \mPhi(\mX) := \left[\begin{matrix}
             \phi_1(\vx_1) &... &\phi_m(\vx_1) \\ 
             \vdots                                  &\ddots &\vdots \\
             \phi_1(\vx_m) &... &\phi_m(\vx_m) 
             \end{matrix}\right].
    \end{equation*}
\end{lemma}

\begin{proof}
Clearly, $\det(\mPhi(\cdot)):\sR^{d\times m}\to\sR$ is an analytic function over $\sR^{d\times m}$. In addition, because $\{\phi_i\}_{i=1}^m$ are linearly independent, there exists $\mX\in\sR^{d\times m}$ such that $\det(\mPhi(\cdot))\neq 0$, i.e., $\det(\mPhi(\cdot))$ is a non-zero analytic function. Therefore, $\rrank(\mPhi(\mX))= m$ a.e. with respect to $\fL^{d\times m}$  Lebesgue measure.
\end{proof}

\begin{cor}\label{cor:zero-measure}
Given $m$ linearly independent analytic functions $\phi_1(\vx),\cdots,\phi_m(\vx)$ with $\phi_i:\sR^d\to\sR$ for all $i\in[m]$ and $\rdim(\rspan(\{\phi_i(\cdot)\}_{i=1}^m))=r$, $\rrank(\mPhi(\mX))= \min\{n,r\}$ a.e. with respect to $\fL^{d\times m}$ Lebesgue measure.
\end{cor}
\begin{proof}
It is obvious that $\rrank(\mPhi(\mX))\leqslant \min\{n,r\}$.
For $n\leqslant r$, we can always pick $n$ independent functions from $\{\phi_i(\cdot)\}_{i=1}^m$. By Lemma \ref{lem:zero-measure}, $\mPhi(\mX)$ has a rank-$n$ submatrix of $\mPhi(\mX)$ a.e. with respect to Lebesgue measure.
For $n>r$, we have that the submatrix of the first $r$ rows of $\mPhi(\mX)$ has rank $r$ a.e. by Lemma \ref{lem:zero-measure}. Therefore, $\rrank(\mPhi(\mX))= \min\{n,r\}$ a.e. with respect to $\fL^{d\times m}$ Lebesgue measure.
\end{proof}

\begin{thm}[phase transition of linear stability for recovery]\label{appthm:phase-transition}
Given any analytic model $f_{\vtheta}$, for any target function $f^*\in\fF_{f_{\vtheta}}$ and $n$ generic training data $S=\{(\vx_i,f^*(\vx_i))\}_{i=1}^n$, \\
(i) \textbf{Strictly under-determined regime:} if $n<\rrank_{f_{\vtheta}}(f^*)$, then $f^*$ is not linearly stable;\\
(ii) \textbf{Quasi-determined regime:} if $n\geqslant\rrank_{f_{\vtheta}}(f^*)$, then $f^*$ is linearly stable almost everywhere with respect to $S$.\\
\end{thm}

\begin{proof}
(i) For any $\vtheta^*\in\fM_{f^*}$, we have $\rrank_S(\vtheta^*) \leqslant n < \rrank_{f_{\vtheta}}(f^*)\leqslant \rrank_{f_{\vtheta}}(\vtheta^*)$. Therefore the linear stability condition cannot be satisfied, i.e., $f^*$ is not linearly stable.

(ii) Given any $\vtheta^*\in\fM_{f^*}$ with $\rrank(\vtheta^*)=\rrank_{f_{\vtheta}}(f^*)$, by Corollary \ref{cor:zero-measure}, $\rrank (\nabla_{\vtheta}f(\mX;\vtheta^*)^\TT) = \min\{n, \rrank_{f_{\vtheta}}(\vtheta^*)\}$ almost everywhere. Because $n\geqslant\rrank_{f_{\vtheta}}(f^*)$, we have $\rrank_S(\vtheta^*) = \rrank_{f_{\vtheta}}(\vtheta^*)$ almost everywhere. By the linear stability condition Lemma \ref{thm:LSC}, $f^*$ is linearly stable almost everywhere.
\end{proof}

\begin{cor}[implicit bias of linear stability hypothesis]\label{appcor:implicitbias}
Given any model $f_{\vtheta}$ and training dataset $S=\{(\vx_i,y_i)\}_{i=1}^n$, if an interpolation $f'\in\fF_{f_{\vtheta}}$ is linearly stable, then $\rrank_{f_{\vtheta}}(f')\leqslant n$.
\end{cor}
\begin{proof}
By the linear stability condition Lemma \ref{thm:LSC}, there exists $\vtheta'\in\fM_{f'}$ such that $\rrank_S(\vtheta') = \rrank_{f_{\vtheta}} (\vtheta')$. Because $\rrank_S(\vtheta')\leqslant n$, we have 
$\rrank_{f_{\vtheta}} (\vtheta')\leqslant n$. Then $\rrank_{f_{\vtheta}}(f')\leqslant\rrank_{f_{\vtheta}} (\vtheta') \leqslant n$.
\end{proof} 

\section{Details of experiments}\label{appsec:exp}
For Fig. \ref{fig:matrix_com}, the target matrices we use are as follows:
\begin{align*}
&\mM_{1}^*=\left[\begin{array}{cccc}
1 &0.3 &0.7 & -0.4   \\
2 &   0.6 & 1.4 &-0.8   \\
4 &   1.2&  2.8& -1.6   \\
 7 &  2.1 & 4.9 & -2.8   \\
\end{array}\right],    
&&\mM_{2}^*=\left[\begin{array}{cccc}
4 &  0.6 & 1.8 & 0.8   \\
6 &  0.9 & 2.7 & 1.2   \\
8 &  1.2 & 3.6 & 1.6   \\
18  & 2.7 & 8.1 & 3.6   \\
\end{array}\right],\\
&\mM_{3}^*=\left[\begin{array}{cccc}
-1.8 & 2.4 & 7.7 & -5.3   \\
0.4 & 1.8 & 5.4 & -3.6   \\
3.2 & 1.8 & 4.8 & -3.   \\
6.6 & 2.4 & 5.9 & -3.5   \\
\end{array}\right],  
&& \mM_{4}^*=\left[\begin{array}{cccc}
7.6 & 3.3 & 19.8 & -7.3   \\
7.6 & 2.1 & 10.7 & -2.4   \\
8.8 & 1.8 & 7.6 & -0.2   \\
19.2 & 3.6 & 14.1 &  0.9   \\
\end{array}\right],\\   
&\mM_{5}^*=\left[\begin{array}{cccc}
-1.8 & 2.4 & 7.7 & -5.3  \\
 0.4 & 1.8 & 5.4 & -3.6  \\
3.2 & 1.8 & 4.8 & -3   \\
6.6 & 2.4 & 5.9 & -3.5   \\
\end{array}\right],
&&\mM_{6}^*=\left[\begin{array}{cccc}
8.5 & 9.3 & 22.5 & -6.1  \\
8.2 & 6.1 & 12.5 & -1.6   \\
11.5 & 19.8 & 15.7 &  3.4   \\
20.4 & 11.6 & 17.7 & 2.5   \\
\end{array}\right],\\
&\mM_{7}^*=\left[\begin{array}{cccc}
3.6 & -1.2 & 8.1 & -3.5   \\
8.1 & -3.5 & 3.6 & -1.2   \\
 9.1 & -1.7 & 11.4 & -0.6   \\
11.4 & -0.6 & 9.1 & -1.7   \\
\end{array}\right],
&&\mM_{8}^*=\left[\begin{array}{cccc}
12.1 & 17.3 & 24.1 & -4.9   \\
16.3 & 24.1 & 16.1 & 1.1   \\
14.2 & 25.8 & 16.9 & 4.3   \\
22.2 & 15.6 & 18.5 & 3.1   \\
\end{array}\right].
\end{align*}

For Fig. \ref{fig:diff_ini}, the target matrix we use is $\mM_{1}^*$ as defined above.

For Fig. \ref{fig:matrix_com2}, the target matrix we use is $\mM_{2}^*$ as defined above. The sampling sequences we use are listed as follows for each row in Fig. \ref{fig:matrix_com2}, respectively:
$$
\{(3,1),(4,3),(2,1),(1,3),(2,4),(4,1),(1,1),(1,2),(4,2),(4,4),(3,2),(3,4),(3,3),(2,2),(2,3),(1,4)\}, 
$$
$$
\{(3,4),(2,1),(2,3),(4,3),(4,1),(4,4),(1,1),(3,3),(1,2),(1,4),(1,3),(2,4),(3,2),(2,2),(3,1),(4,2)\},
$$
$$
\{(2,4),(3,3),(3,1),(4,4),(4,3),(3,4),(1,3),(1,4),(2,3),(3,3),(1,1),(1,2),(4,2),(2,2),(2,1),(4,1)\},
$$
$$
\{(4,4),(2,3),(4,2),(1,2),(1,4),(3,2),(4,1),(3,1),(1,1),(3,4),(1,3),(2,2),(2,4),(2,1),(3,3),(4,3)\},
$$
$$
\{(2,4),(3,4),(4,1),(1,2),(2,2),(4,4),(1,1),(3,1),(3,2),(4,2),(2,1),(1,3),(4,3),(3,3),(2,3),(1,4)\},
$$
$$
\{(4,3),(4,4),(2,1),(3,4),(3,3),(3,1),(2,3),(1,1),(4,1),(2,4),(1,4),(1,3),(1,2),(2,2),(3,2),(4,2)\}.
$$

For Fig. \ref{fig:eig_dis}, the target matrix we use is $\mM_{2}^*$ as defined above. For three sets of experiments with $7,12,15$ training samples, we sample the following sets of indices of the target matrix, respectively:
$$
\{(1,1),(1,2),(1,3),(1,4),(2,2),(3,3),(4,4)\},
$$
$$
\{(1,1),(1,2),(1,3),(1,4),(2,1),(2,2),(2,3),(2,4),(3,3),(3,4),(4,3),(4,4)\},
$$
$$
\{(1,1),(1,2),(1,3),(1,4),(2,1),(2,2),(2,3),(2,4),(3,1),(3,2),(3,3),(3,4),(4,2),(4,3),(4,4)\}.
$$

For Fig. \ref{fig: network_stab}, we consider the following target function:

$$\boldsymbol{f}_{\boldsymbol{\theta}}(\boldsymbol{x}) = \boldsymbol{W}^{[2]}\sigma(\boldsymbol{W}^{[1]}\boldsymbol{x}),$$ 

where $\boldsymbol{W}^{[1]}=\left[\begin{array}{ccccc}
0.6 &0.8 &1 & 0 &0  \\
0 &0.6 &0.8 &1 & 0  \\
0 &0 &0.6 &0.8 &1  \\
\end{array}\right]$, $\boldsymbol{W}^{[2]}=[1,1,1]$. 

For the training dataset and the test dataset, we construct the input data through the standard normal distribution and obtain the output values from the target function. The size of the training dataset varies whereas the size of the test dataset is fixed to $1000$. The learning rate for the experiments in each setup is fine-tuned from $0.05$ to $0.5$ for a better generalization performance.

\end{document}